\documentclass[runningheads]{llncs}
\usepackage{graphicx}
\graphicspath{ {figures/} }

\usepackage{tikz}
\usepackage{comment}
\usepackage{amsmath,amssymb} 
\usepackage{amsfonts}
\usepackage{bm}
\usepackage{color}
\usepackage{orcidlink}
\usepackage[misc]{ifsym}

\usepackage{listings}
\usepackage{xcolor}
\definecolor{codegreen}{rgb}{0,0.6,0}
\definecolor{codegray}{rgb}{0.5,0.5,0.5}
\definecolor{codepurple}{rgb}{0.58,0,0.82}
\definecolor{backcolour}{rgb}{0.97,0.97,0.97}
\lstdefinestyle{mystyle}{
    backgroundcolor=\color{backcolour},   
    commentstyle=\color{codegreen},
    keywordstyle=\color{magenta},
    numberstyle=\tiny\color{codegray},
    stringstyle=\color{codepurple},
    basicstyle=\ttfamily\footnotesize,
    breakatwhitespace=false,         
    breaklines=true,                 
    captionpos=b,                    
    keepspaces=true,                 
    numbers=left,                    
    numbersep=5pt,                  
    showspaces=false,                
    showstringspaces=false,
    showtabs=false,                  
    tabsize=2
}
\newcommand{\etal}{\textit{et al}.}
\newcommand{\ie}{\textit{i}.\textit{e}.}
\newcommand{\eg}{\textit{e}.\textit{g}.}
\newcommand{\etc}{\textit{etc}.}

\usepackage[accsupp]{axessibility}  

\begin{document}
\pagestyle{headings}
\mainmatter
\def\ECCVSubNumber{636}  

\title{Sem2NeRF: Converting Single-View \\ Semantic Masks to Neural Radiance Fields} 

\titlerunning{Sem2NeRF}
%
\author{Yuedong Chen\inst{1}\orcidlink{0000-0003-0943-1512}\textsuperscript{\Letter} \and
Qianyi Wu\inst{1}\orcidlink{0000-0001-8764-6178} \and
Chuanxia Zheng\inst{1}\orcidlink{0000-0002-3584-9640} \and \\
Tat-Jen Cham\inst{2}\orcidlink{0000-0001-5264-2572} \and
Jianfei Cai\inst{1}\orcidlink{0000-0002-9444-3763}
}
\authorrunning{Y. Chen et al.}
%
\institute{Monash University, Australia \\
\email{\{yuedong.chen, qianyi.wu, jianfei.cai\}@monash.edu}
\and
Nanyang Technological University, Singapore \\
\email{chuanxia001@e.ntu.edu.sg, astjcham@ntu.edu.sg}}
\maketitle

\begin{abstract}

Image translation and manipulation have gain increasing attention along with the rapid development of deep generative models. Although existing approaches have brought impressive results, they mainly operated in 2D space. 
In light of recent advances in NeRF-based 3D-aware generative models,
we introduce a new task, Semantic-to-NeRF translation, that aims to reconstruct a 3D scene modelled by NeRF, conditioned on one single-view semantic mask as input. To kick-off this novel task, we propose the Sem2NeRF framework.
In particular, Sem2NeRF addresses the highly challenging task by encoding the semantic mask into the latent code that controls the 3D scene representation of a pre-trained decoder. To further improve the accuracy of the mapping, we integrate a new region-aware learning strategy into the design of both the encoder and the decoder. We verify the efficacy of the proposed Sem2NeRF and demonstrate that it outperforms several strong baselines on two benchmark datasets. Code and video are available at \url{https://donydchen.github.io/sem2nerf/}.

\keywords{NeRF-based generation, conditional generative model, 3D deep learning, neural radiance fields, image-to-image translation}
\end{abstract}

\section{Introduction} \label{sec:intro}

Controllable image generation, translation, and manipulation have seen rapid advances in the last few years along with the emergence of Generative Adversarial Networks (GANs)~\cite{goodfellow2014generative}. Current systems are able to freely change the image appearance through referenced images~\cite{johnson2016perceptual,zhu2017unpaired,isola2017image}, modify scene content via semantic masks~\cite{wang2018high,park2019semantic,ling2021editgan}, and even accurately manipulate various attributes in feature space~\cite{karras2019style,wu2021stylespace,wu2021stylealign}. Despite impressive performance and wide applicability, these systems are mainly focused on 2D images, without directly considering the 3D nature of the world and the objects within. 

Concurrently, significant progress has been made for 3D generation by using deep generative networks~\cite{goodfellow2014generative,kingma2014adam}. Methods were developed for different 3D shape representations, including voxels~\cite{wu2016learning}, point clouds~\cite{luo2021diffusion}, and meshes~\cite{goel2020shape}. More recently, Neural Radiance Fields (NeRF)~\cite{mildenhall2020nerf} has been a new paradigm for 3D representation, providing accurate 3D shape and view-dependent appearance simultaneously. Based on this new representation, seminal 3D generation approaches~\cite{schwarz2020graf,niemeyer2021giraffe,chan2021pi,gu2021stylenerf,chan2021efficient} have been proposed that aim to generate photorealistic images from a given distribution in a 3D-aware and view-consistent manner. However, these techniques are primarily developed purely for high-quality 3D generation, leaving controllable 3D manipulation and editing unsolved.

\begin{figure}[t!]
    \centering
    \includegraphics[width=0.99\textwidth]{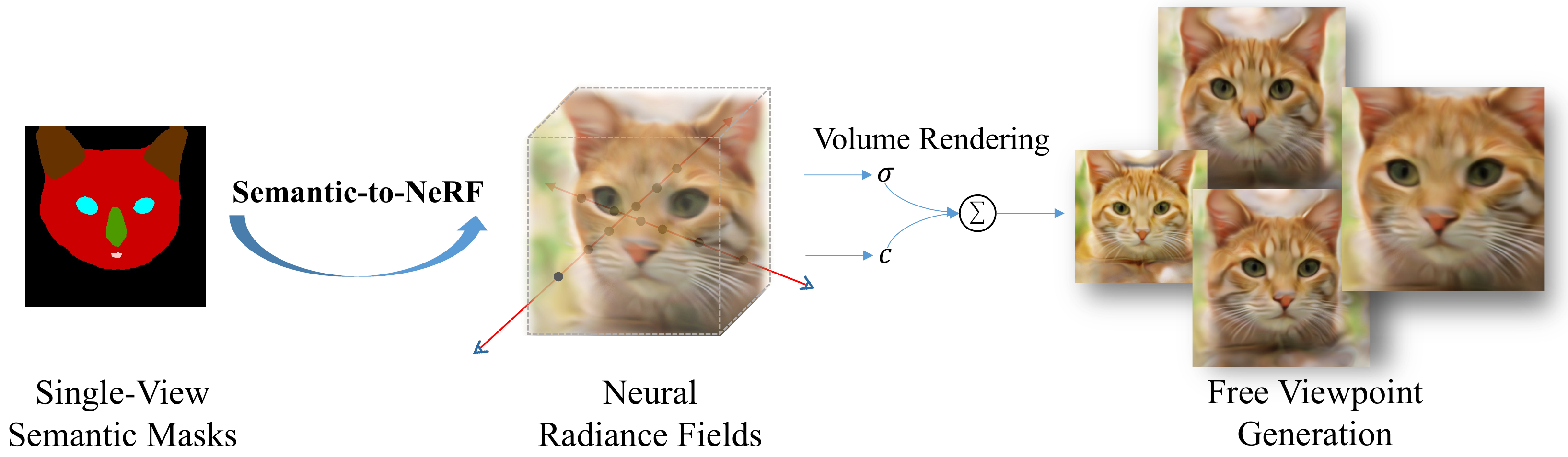}
    \caption{Illustration of the Semantic-to-NeRF translation task, which aims to achieve free-viewpoint image generation by taking only a single-view semantic mask as input}
    \label{fig:semantic_to_nerf}
\end{figure}

It would be a dramatic enhancement if we can \emph{freely manipulate and edit an object's content and appearance in 3D space, while only leveraging easily obtained 2D input information}. In this paper, we take an initial step toward this grand goal by introducing a new task, termed \textbf{Semantic-to-NeRF translation}, analogous to a 2D Semantic-to-Image translation task but operating on 3D space. Specifically, Semantic-to-NeRF translation (see Fig.~\ref{fig:semantic_to_nerf}) takes as input a single-view 2D semantic mask, yet output a NeRF-based 3D representation that can be used to render photorealistic images in a 3D-aware view-consistent manner. More importantly, it allows free editing of the object's content and appearance in 3D space, by modifying the content only via a single-view 2D semantic mask.

However, generating 3D structure from a single 2D image is already an ill-posed problem, and it will be even more so from a single 2D semantic mask. There are also two other major issues in this novel task:
\begin{enumerate}
    \item \emph{Large information gap between 3D structure and 2D semantics.} A single-view 2D semantic mask neither holds any 3D shape or surface information, nor provides much guidance for plausible appearances, making it tough to generate a neural radiance field with comprehensive details.
    \item \emph{Imbalanced semantic distribution.} Since semantic classes tend to be area-imbalanced within an image, \eg~ eyes occupy less than $1\%$ of a face while hair can take up larger than $40\%$, existing CNN-based networks may over-attend to larger semantic regions, while discounting smaller semantic regions that may be perceptually more salient. This will result in poor controllable editing in 3D space when we alter small semantic regions.
\end{enumerate}
To mitigate these issues, we propose a novel framework, \textbf{Sem2NeRF}, that builds on NeRF~\cite{mildenhall2020nerf} for 3D representation, by augmenting it with a semantic translation branch that conditionally generates high-quality 3D-consistent images. In particular, the framework is based on an encoder-decoder architecture that converts a singe-view 2D semantic mask to an embedded code, and then transfers it to a NeRF representation for rendering  3D-consistent images. 

Our broad idea here is that, instead of directly learning to predict 3D structure from degenerate single-view 2D semantic masks, \emph{the network can alternatively learn the 3D shape and appearance representation from large numbers of unstructured 2D RGB images}. This has achieved significant advances in NeRF-based generator~\cite{schwarz2020graf,niemeyer2021giraffe,chan2021pi,gu2021stylenerf,chan2021efficient}, which transforms a random vector to a NeRF representation. In short, our scenario is thus: we have a well-trained 3D generator, but we aim to further control the generated content and appearance easily. The main idea is then to \emph{learn a good mapping network} (like current methods for 2D GAN inversion \cite{richardson2021encoding,song2021agilegan}) \emph{that can encode the semantic mask into the somewhat smaller latent space domain for 3D controllable translation and manipulation.} As for the second issue, we intriguingly discover that a \emph{region-aware learning strategy} is of vital importance. We therefore aim to tame an encoder that is sensitive to image patches, and adopt a region-based sampling pattern for the decoder. Furthermore, augmenting the input semantic masks with extracted contours and distance field representations~\cite{chen2018sketchygan} also considerably helps to highlight the intended semantic changes, making them more easily perceptible.

Following the above analysis, we build our Sem2NeRF framework upon the Swin Transformer encoder~\cite{liu2021swin} and the pre-trained $\pi$-GAN decoder~\cite{chan2021pi}. To kick off the single-view Semantic-to-NeRF translation task, we pinpoint two suitable yet challenging datasets, including CelebAMask-HQ~\cite{lee2020maskgan} and CatMask, where the latter contains cat faces rendered using $\pi$-GAN and labelled with 6-class semantic masks using DatasetGAN~\cite{zhang2021datasetgan}. We showcase the superiority of our model over several strong baselines by considering SofGAN~\cite{chen2022sofgan}, pix2pixHD~\cite{wang2018high} with GAN-inversion~\cite{karras2020analyzing}, and pSp~\cite{richardson2021encoding}. 
Our contributions are three-fold:
\begin{itemize}
  \item We introduce a novel and challenging task, Semantic-to-NeRF translation, which converts a single-view 2D semantic mask to a 3D scene modelled by neural radiance fields.
  \item With the insight of needing a region-aware learning strategy, we propose a novel framework, Sem2NeRF, which is capable of achieving 3D-consistent free viewpoint image generation, semantic editing and multi-model synthesis, by taking as input only one single-view semantic mask of a specific category, \eg, human face, cat face.
  \item We validate our insight regarding our region-aware learning strategy and the efficacy of Sem2NeRF via extensive ablation studies, and demonstrate that Sem2NeRF outperforms strong baselines on two challenging datasets. 
\end{itemize}

\section{Related Work}

\textbf{NeRF and Generative NeRF.}
Starting as an approach focused on modelling a single static scene, NeRF~\cite{mildenhall2020nerf} had seen rapid development in different aspects. Several approaches managed to reduce the training~\cite{sun2021direct} and inference time~\cite{liu2020neural,lombardi2021mixture}, while others improved visual quality~\cite{barron2021mip}. Besides, it had also been extended in other ways, \eg, dynamic scene~\cite{pumarola2021d}, compositional scene~\cite{wu2022objectsdf}, pose estimation~\cite{yen2021inerf}, portrait generation~\cite{liu2022semantic}, semantic segmentation~\cite{zhi2021place}. 

Follow-up works that integrated NeRF with generative models were most relevant to ours. Schwarz \etal~\cite{schwarz2020graf} proposed to learn a NeRF distribution by conditioning the input point positions with a sampled random vector.
Niemeyer \etal~\cite{niemeyer2021giraffe} enabled multi-object generation by representing the whole scenes as a composition of different components. To improve the visual quality, $\pi$-GAN~\cite{chan2021pi} adopted a SIREN-based~\cite{sitzmann2020implicit} network structure with FiLM~\cite{perez2018film} conditioning. StyleNeRF~\cite{gu2021stylenerf} turned to embedding the volume rendering technique into StyleGAN~\cite{karras2020analyzing}. 
More recently, VolumeGAN~\cite{xu20213d} relied on separately learning structure and texture features. MVCGAN~\cite{zhang2022multi} leveraged the underlying 3D geometry information. EG3D~\cite{chan2021efficient} proposed an efficient tri-plane hybrid 3D representation.

Our work belongs to the class of generative models, but unlike all existing methods that aimed to create a \emph{random} scene, we aim to generate a \emph{specific} scene that is conditioned by a given single-view semantic mask. Although there are concurrent works, \eg, 3D-SGAN~\cite{zhang20213d}, FENeRF~\cite{sun2022fenerf}, exploring the similar condition settings, most of them purely focus on improving the quality of the generated images, while resort to existing GAN inversion~\cite{karras2020analyzing} to do the mapping. In contrast, our work is more focused on improving the mapping from the mask to the NeRF-based scene.

\noindent\textbf{Image-to-Image Translation} is about converting an image from one source representation, \eg, semantic masks, to another target representation, \eg, photorealistic images. Since its introduction~\cite{isola2017image}, progress has been made with regard to better image quality~\cite{wang2018high,chen2021edge}, multi-modal outputs~\cite{zhu2017toward,zheng2019pluralistic,choi2020stargan}, unsupervised learning~\cite{zhu2017unpaired,lira2020ganhopper}, \etc.  More recently,
there is a new trend~\cite{richardson2021encoding,shi2022semanticstylegan,xu2022transeditor} of tackling this task by editing the latent space of a pre-trained generator, \eg, StyleGAN.

In contrast to all mentioned work that aimed to map a semantic mask to an image, ours is focused on mapping to a 3D scene.
We also notice that there are some recent approaches targeted at converting semantic masks to 3D scenes. Huang \etal~\cite{huang2020semantic} introduced rendering novel-view photorealistic images from a given semantic mask, by first applying semantic-to-image translation~\cite{park2019semantic}, then converting the single-view image to a 3D scene modelled by multiplane images (MPI)~\cite{zhou2018stereo}.
Hao \etal~\cite{hao2021gancraft} proposed to learn a mapping from a semantically-labelled 3D block world to a NeRF-based 3D scene, using a scene-specific setting. Chen \etal~\cite{chen2022sofgan} introduced a 3D-aware portrait generator by first mapping the given latent code to a semantic occupancy field (SOF)~\cite{chen2019learning} for rendering novel view semantic masks, followed by applying image-to-image translation. 

Unlike all mentioned attempts on learning semantic to 3D scene mappings, ours is the first to introduce the single-view semantic to NeRF translation task. Our work differs from theirs in: 1) We do not rely on any separate image-to-image translation stage, resulting in better multi-view consistency; 2) We do not require multi-view semantic masks for both training and testing phases, easing the data collection effort; 3) We pinpoint a solution for creating pseudo labels and demonstrate reasonable results beyond the human face domain.

\begin{figure}[t!]
    \centering
    \includegraphics[width=0.99\textwidth]{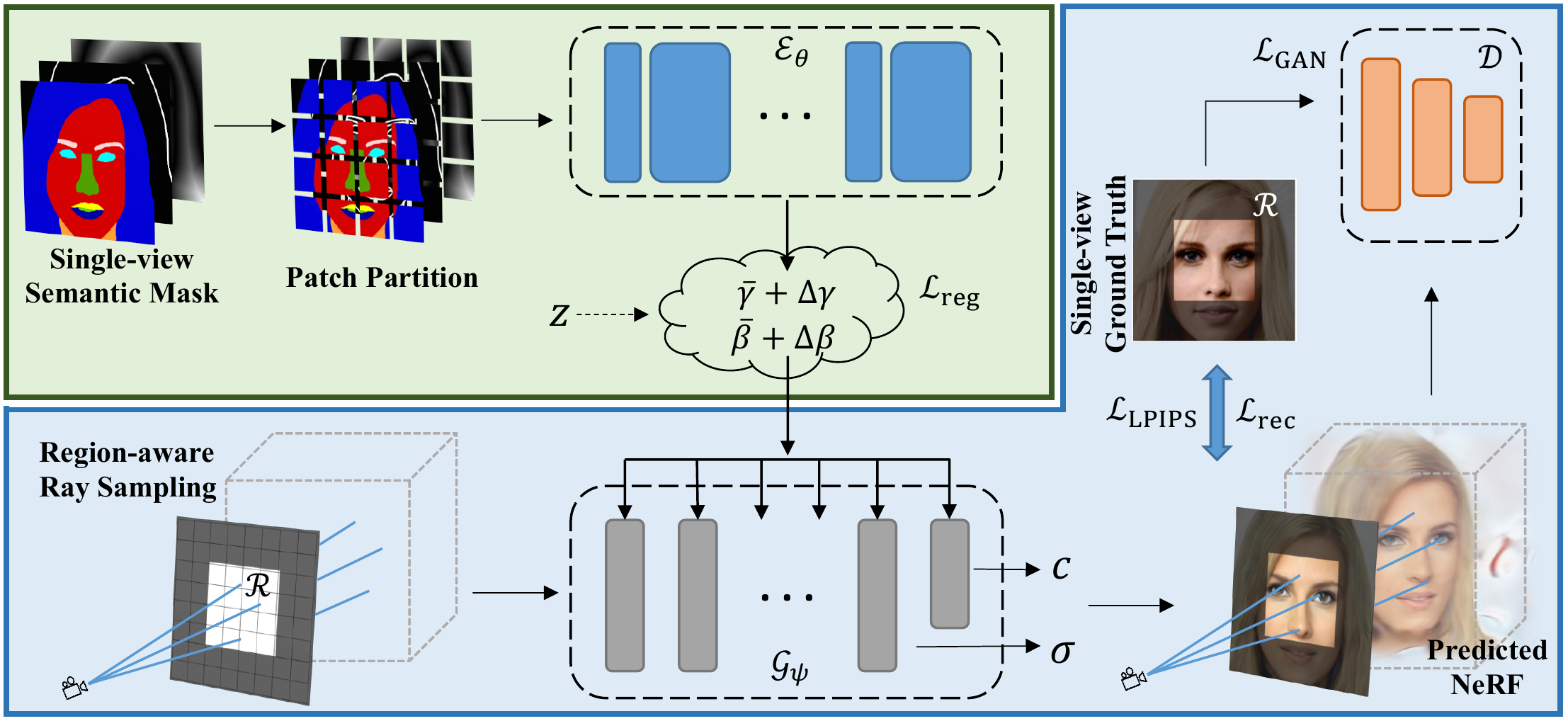}
    \caption{Architecture of the Sem2NeRF framework. It aims to convert a single-view semantic mask to a 3D scene represented by NeRF. Specifically, a given semantic mask will be partitioned into patches, which will be further encoded by a patch-based encoder $\mathcal{E}_\theta$ into a latent style code $(\gamma,\beta)$ of a pre-trained NeRF-based 3D generator $\mathcal{G}_\psi$. A region $\mathcal{R}$ will be randomly sampled to enforce awareness of differences among regions. And an optional latent vector $\mathbf{z}$ is included to enable multi-modal synthesis
    }
    \label{fig:framework_archi}
\end{figure}

\section{Methodology}

As shown in Fig.~\ref{fig:semantic_to_nerf}, our main goal is to train a Semantic-to-NeRF translation network $\Phi_{s\to\mathcal{V}}$, such that when presented with a single-view 2D semantic mask $\mathbf{s}$, it generates the corresponding NeRF representation $\mathcal{V}$, which can then be used to render realistic 3D-consistent images. This task is conceptually similar to the conventional semantic-to-image setting, except that here we opt to go beyond 2D image translation, and deal with the novel \emph{controllable 3D translation.} More importantly, we can freely change the 3D content by \emph{simply modifying the corresponding content in a single-view 2D semantic mask}.

In order to learn such a framework without enough supervision for arbitrary view appearances, we observed that 3D information can be learned from large image collections~\cite{kanazawa2018learning,chan2021pi,chan2021efficient}. Therefore, our \emph{key motivational insight} is this: instead of directly training $\Phi_{s\to\mathcal{V}}$ using \emph{single-view} semantic-image pairs $(\mathbf{s},\mathbf{I})$ (like current methods for 2D semantic-to-image translation~\cite{wang2018high,park2019semantic}), we will train it as a two-stage pipeline shown in Fig.~\ref{fig:framework_archi}. Here, (A) we utilize a pre-trained 3D generator (lower portion $\mathcal{G}_\psi$) that learns 3D shape and appearance information from a large set of collected images; 
(B) 
we pose this challenging task as a \emph{3D inversion} problem, where our main target is to design a front-end encoder (upper portion $\mathcal{E}_\theta$) that maps the semantic mask into the generator latent space accurately. 

The two training stages are executed independently and can be separately implemented with different frameworks. There are at least two unique benefits of breaking down the entire controllable 3D translation into two-stages: 1) The training does \emph{not} require copious views of semantic-image pairs for each instance, which are difficult to collect, or even impossible in some scenarios; 2) The compartmentalization of the 3D generator and the 2D encoder allows greater agility, where the 3D information can be previously learned on various tasks with a large collection of images and then be freely plugged into the 3D inversion pipeline.

\subsection{3D Decoder with Region-Aware Ray Sampling}

\paragraph{\textbf{Preliminaries on NeRF.}} We first provide some preliminaries on NeRF before discussing how we exploit it for Sem2NeRF. NeRF~\cite{mildenhall2020nerf} is one kind of implicit functions that represents a continuous 3D scene, which has achieved great successes in modeling 3D shape and appearance. A NeRF is a neural network that maps a 3D location $\mathbf{x}\in\mathbb{R}^3$ and a viewing direction $\mathbf{d}\in\mathbb{S}^2$ to a spatially varying volume density $\sigma$ and a view-dependent emitted color $\mathbf{c} = (r, g, b)$. NeRFs trained on natural images are able to continuously render realistic images at arbitrary views. In particular, it requires to use the volume rendering~\cite{levoy1990efficient}, 
which computes the following integral to obtain the color of a pixel:
\begin{equation} \label{eq:volrender}
C(\mathbf{r}) = \int_{t_n}^{t_f}T(t)\sigma(\mathbf{r}(t))\mathbf{c}(\mathbf{r}(t), \mathbf{d})dt, \text{where}~T(t) = \text{exp}(-\int_{t_n}^{t}\sigma(\mathbf{r}(s))ds),
\end{equation}
where $\bm{r}(t) = \bm{o} + t\bm{d}$ is the ray casting from the virtual camera located at $\bm{o}$, bounded by near $t_n$ and far $t_f$, and $T(t)$ represents the accumulated transmittance of the ray traveling from $t_n$ to $t$. The integral $C(\mathbf{r})$ is further implemented with a hierarchical volume sampling strategy~\cite{levoy1990efficient,mildenhall2020nerf}, resulting in the optimization of a ``coarse'' network followed by a ``fine'' network.

\paragraph{\textbf{NeRF-based Generator.}} Our work is mainly based on a representative NeRF-based generator, $\pi$-GAN~\cite{chan2021pi}, which learns 3D representation using only 2D supervision. Inspired by StyleGAN2~\cite{karras2020analyzing}, the architecture of $\pi$-GAN is mainly composed of two parts, a mapping network $\mathcal{F}:\mathcal{Z}\to\mathcal{W}$ that maps a latent vector $z$ in the input latent space $\mathcal{Z}$ to an intermediate latent vector $w\to\mathcal{W}$, and a SIREN-based~\cite{sitzmann2020implicit} synthesis network that maps $w$ to the NeRF representation $\mathcal{V}$ that supports rendering 3D-consistent images from arbitrary camera poses.

Our Sem2NeRF framework can use various NeRF-based generators. Here, we choose $\pi$-GAN as the main decoder in our architecture for two main reasons. Firstly, among all \emph{published} works related to NeRF-based generators, $\pi$-GAN achieves state-of-the-art performance in terms of rendered image quality and their underlying 3D consistency. Secondly and more importantly, similar to StyleGAN, the FiLM~\cite{perez2018film} conditioning used by $\pi$-GAN enables layer-wise control over the decoder and the mapping network decouples some high-level attributes, making it easier to perform \emph{3D inversion} on top of NeRF, \ie, searching for the desired latent code $w$ that best reconstructs an ideal target. The similar observation has been previously explored in the latest 2D GAN inversion \cite{collins2020editing,karras2020analyzing,richardson2021encoding}.

\paragraph{\textbf{Region-Aware Ray Sampling.}} While $\pi$-GAN already provides high-quality view-consistent rendered images, our main goal is to accurately restore the NeRF from a single-view semantic mask, and even freely edit the 3D content via such a map. To achieve this, \emph{the network should be sensitive to local small modifications.} However, this is not supported in the original $\pi$-GAN, which is trained on each entire image with a global perception. It stores scene-specific information in a latent code, which is shared across all points that are bounded by the rendering volume. As a result, a small change in an original latent code will easily cause a global modification in generation. This may not impact pure 3D generation, for which only the quality of global shape and appearance is paramount, but it has a large negative effect on recreating a 3D representation that accurately matches the corresponding semantic mask.

To mitigate this issue, we adopt a region-based ray sampling pattern~\cite{schwarz2020graf,liu2022semantic} in the $\pi$-GAN decoder, that \emph{attempts to encourage latent codes to represent local regions at different scales and locations}. Suppose the rendered image $\mathbf{I}$ with a target size $h\times w$, a local region $\mathcal{R}$ used for training is randomly sampled as
\begin{equation}
\mathcal{R}(\alpha, (\Delta h, \Delta w)) = \left \{ (\alpha h + \Delta h, \alpha w + \Delta w)  \right \},
\label{eq:ray_sample}
\end{equation}
where $(\alpha h + \Delta h, \alpha w + \Delta w)$ denotes the sampling coordinates of rays, with $\alpha \in (0, 1]$ being the scaling factor and $(\Delta h \in [0, (1 - \alpha)h], \Delta w \in [0, (1 - \alpha)w])$ being the translation factor. To obtain such training pairs between the NeRF rendered output and the local ground truth, we sample the original whole image using the same region coordinates $\mathcal{R}$ with bilinear interpolation. This strategy leads to large improvements on conditional generation as shown in the experiments.

\subsection{3D Inversion Using Region-Aware 2D Encoder}\label{sec:method_enc}

\paragraph{\textbf{3D Inversion.}} To inversely map a semantic mask $\mathbf{s}$ into the $\mathcal{W}$ latent space of the 3D generator $\mathcal{G}_\psi$ by an encoder  $\mathcal{E}_\theta$, with respective parameters $\psi$ and $\theta$, we train $\mathcal{E}_\theta$ to minimize the reconstruction error between ground truth image $\mathbf{I}$ and output  $\hat{\mathbf{I}}$. Specifically, Semantic-to-NeRF translation represents the mapping 
\begin{equation} \label{eq:s2n_mapping}
\Phi_{\mathbf{s}\to\mathcal{V}}(\mathbf{x},\mathbf{d},\mathbf{z};\mathbf{s})=\mathcal{G}_\psi(\mathbf{x},\mathbf{d},\mathbf{z};\mathcal{E}_\theta(\mathbf{s}))=\mathcal{V}(\sigma,\mathbf{c})
\end{equation}
where $\mathbf{x}, \mathbf{d}$ denotes point position and ray direction, while the derived density $\sigma$ and color $\bm{c}$ can be used to calculate the corresponding pixel value via volume rendering as in Eq.~\eqref{eq:volrender}. For \emph{controllable} 3D generation, $\mathbf{s}$ is the input single-view semantic mask, embedded into $\mathcal{W}$ space to control the generated 3D content, while we also enable multi-modal synthesis by adding another latent vector $\mathbf{z}$ to model the generated appearance. Note that $\mathbf{s}$ only comes in a single view, which is not necessary the same as the output viewing direction. 
In short, \emph{we use only single-view semantic-image pairs $(\mathbf{s}, \mathbf{I})$ for the Sem2NeRF training}, as the 3D view-consistent information has been captured by the \emph{fixed} pre-trained 3D generator $\mathcal{G}_\psi$. Hence, we focus only on training the encoder network $\mathcal{E}_\theta$ to learn the posterior distribution $q(w|\mathbf{s})$ for 3D inversion.

\paragraph{\textbf{Region-Aware 2D Encoder.}} A simple way for 3D inversion is to directly apply an existing 2D GAN inversion framework. 
However, this straightforward idea does \emph{not} work well
as we originally discovered when using the state-of-the-art pSp encoder~\cite{richardson2021encoding} in our setting,
especially for small but perceptually important regions, such as eyes. Our conjecture is that the conventional CNN-based architecture integrates the neighboring information via overaggressive filtering, resulting in heavy loss of small details \cite{zhang2019making}. 

To mitigate this issue, we also deploy a region-aware learning strategy in the 2D encoder, which is inspired by the latest patch-based methods \cite{dosovitskiy2020image,zheng2021tfill} that capture information in every patch with equal possibility. In other words, when we directly extract features from local patches, it will be \emph{more sensitive to the semantic variation within each patch}, which can ameliorate the problem of imbalanced semantic distribution within an image. In particular, we adopt the Swin Transformer \cite{liu2021swin} as the encoder architecture. To embed the semantic mask $\mathbf{s}$ into the $\mathcal{W}$ latent space of the pre-trained 3D generator, we replace the final classification output size with the size of the latent vectors $\mathbf{w}$. Besides, 
to further stabilize the inversion training, we take inspiration from the truncation trick~\cite{karras2019style,richardson2021encoding} and set the learned latent codes for the pre-trained decoder as

\begin{equation}
    \gamma = \overline{\gamma} + \Delta \gamma, \beta = \overline{\beta} + \Delta \beta,
\end{equation}
where $\gamma$ and $\beta$ represent the embedded vectors for the $\mathcal{W}$ latent space, \ie, frequency and phase shift of $\pi$-GAN, respectively; $\Delta \gamma$ and $\Delta \beta$ are the outputs of the proposed encoder $\mathcal{E}_\theta$, while $\overline{\gamma}$ and $\overline{\beta}$ are the average latent codes extracted by the pre-trained $\pi$-GAN original mapping network $\mathcal{F}:\mathcal{Z}\to\mathcal{W}$.

\paragraph{\textbf{Additional Inputs for the 2D Encoder.}} As mentioned, a semantic mask contains sparse information, where the changing of small regions may be imperceptible to the network, making the semantic-based controllable 3D editing very challenging. Considering that editing a semantic mask only effectively alters the boundaries between different semantic labels, we conjecture that explicitly augmenting the semantic input with \emph{boundary information} will be useful for semantic editing. Therefore, we concatenate the semantic mask input with contours and distance field representations~\cite{chen2018sketchygan} for the region-aware encoder. These additional inputs further improve the semantic editing performance considerably as shown in the experiments. Note that contours and distance field representations are both directly calculated from the semantic masks (refer to Section~\ref{sec:tech_input} for more details), which do \emph{not involve any extra labels}.

\subsection{Training Loss Functions}
During the training phase, we use the single-view semantic mask $\mathbf{s}$, the corresponding viewing direction $d_s$, and the paired ground truth RGB image $\mathbf{I}$. Similar to Semantic-to-Image translation, we start by applying a pixel-level reconstruction loss,
\begin{equation}
    \mathcal{L}_{\text{rec}}(\mathbf{I}, \mathbf{s}, d_s) = \left \| \mathbf{I} - \mathcal{G}_\psi(\mathcal{E}_\theta(\mathbf{s}), d_s) \right \|_2,
\end{equation}
where $\mathcal{E}_\theta(\mathbf{s})$ denotes the latent codes mapped from $\mathbf{s}$ via the region-aware encoder $\mathcal{E}_\theta(\cdot)$, while $\mathcal{G}_\psi(\mathcal{E}_\theta(\mathbf{s}), d_s)$ represents the generated image rendered from direction $d_s$ via the decoder $\mathcal{G}_\psi(\cdot)$. Unless otherwise specified, the aforementioned region-aware sampling strategy is applied to $\mathcal{G}_\psi$ and $\mathbf{I}$ before calculating any losses. 

To further enforce the feature-level similarity between the generated image and the ground truth, the LPIPS loss~\cite{zhang2018unreasonable} is leveraged, 
\begin{equation}
    \mathcal{L}_{\text{LPIPS}}(\mathbf{I}, \mathbf{s}, d_s) = \left \| \mathcal{F}(\mathbf{I}) - \mathcal{F}(\mathcal{G}_\psi(\mathcal{E}_\theta(\mathbf{s}), d_s))) \right \|_2,
\end{equation}
where $\mathcal{F}(\cdot)$ refers to the pre-trained feature extraction network.

Inspired by the truncation trick~\cite{karras2019style,richardson2021encoding}, we further encourage the decoder latent codes $\gamma$, $\beta$ to be close to the average codes $\overline{\gamma}$, $\overline{\beta}$, which is achieved by regularizing the encoder with
\begin{equation}
    \mathcal{L}_{\text{reg}}(\mathbf{s}) = \left \| \mathcal{E}_\theta(\mathbf{s}) \right \|_2.
\end{equation}

To improve image quality, especially for novel views, we further apply a non-saturating GAN loss with R1 regularization~\cite{mescheder2018training}, \begin{equation} \label{eq:gan}
\begin{aligned}
\mathcal{L}_{\text{GAN}}(\mathbf{I}, \mathbf{s}, d) & =  f(\mathcal{D}(\mathcal{G}_\psi(\mathcal{E}_\theta(\mathbf{I}), d))) + f(-\mathcal{D}(\mathbf{I})) + \lambda_{\text{R1}} \left | \nabla \mathcal{D}(\mathbf{I}) \right |^2, \\
    & \text{where}\: f(u)  = -\log(1 + \exp(-u)).
\end{aligned}
\end{equation}
Here $\mathcal{D}(\cdot)$ is a patch discriminator~\cite{isola2017image}, aligned with our region-aware learning strategy for the decoder, and $\lambda_{\text{R1}}$ is a hyperparameter that is set to 10. Note that here the viewing direction $d$ is not required to be the same as the input semantic viewing direction $d_s$, and we randomly sample this viewing direction from a known distribution, \ie~Gaussian, following the settings of $\pi$-GAN~\cite{chan2021pi}.

Finally, the overall training objective for our framework is a weighted combination of the above loss functions as
\begin{equation} \label{eq:total_loss}
    \mathcal{L}_{\text{Sem2NeRF}} = \lambda_{\text{rec}} \mathcal{L}_{\text{rec}} + \lambda_{\text{LPIPS}}\mathcal{L}_{\text{LPIPS}} + \lambda_{\text{reg}}\mathcal{L}_{\text{reg}} + \lambda_{\text{GAN}}\mathcal{L}_{\text{GAN}}.
\end{equation}

\subsection{Model Inference}
For inference, our model takes as input a 2D single-view semantic mask, while $d_s$ is optional, required only when rendering an image with the same viewing direction as the semantic mask. Different from the training phase, during inference the rays are cast to cover the whole image plane, rather than a local region. 

\noindent\textbf{Multi-View Generation.} 
Since the employed decoder is a NeRF-based generator, Sem2NeRF inherently supports novel view generation. Specifically, given a semantic mask $\mathbf{s}$, it will first be mapped as an embedded vector in the $\mathcal{W}$ latent space that controls the ``content'' of the NeRF-based generator, whereupon a novel view image can then be generated by volume rendering the NeRF from an arbitrary viewing direction.  

\noindent\textbf{Multi-Modal Synthesis.} Similar to the diversified mapping in semantic-to-image~\cite{park2019semantic}, ideally a single semantic mask should be translated into multiple NeRFs consistent to it. Our Sem2NeRF framework inherently supports multi-modal synthesis in inference due to the usage of FiLM~\cite{perez2018film} conditioning on $\pi$-GAN, without requiring any special customization in training. In practice, we additionally pass a random-sampled vector to the pre-trained $\pi$-GAN noise mapping module to obtain corresponding latent style codes $\bm{z}$. Style mixing~\cite{richardson2021encoding,karras2019style} is then performed between $\bm{z}$ and $\mathcal{E}_\theta(\mathbf{s})$ to yield multi-modal outcomes.

\section{Experiments}

\subsection{Settings}\label{sec:exp_set}

\textbf{Datasets.} To achieve Semantic-to-NeRF translation, we assume the training data to have single-view registered semantic masks and images, with the corresponding viewing directions. Two datasets were used for evaluation in our experiments.
\textbf{CelebAMask-HQ~\cite{lee2020maskgan}} contains images from CelebA-HQ~\cite{liu2015deep,karras2017progressive}, manually-labelled 19-class semantic masks, 
and head poses. We merged the left-right labels of symmetric parts, \ie, eyes, eyebrows and ears, into one label per part. The dataset was randomly partitioned into training set with 28,000 samples and test set with 2,000 samples.
\textbf{CatMask} is built using $\pi$-GAN and DatasetGAN~\cite{zhang2021datasetgan} to further demonstrate the potential of  Semantic-to-NeRF task and  Sem2NeRF. Technical details are elaborated in Section~\ref{sec:tech_cat}. 

\noindent\textbf{Baselines.} We identified the following three methods as baselines for comparison in our introduced Semantic-to-NeRF task.
\textbf{SofGAN~\cite{chen2022sofgan}} is an image translation approach. For a given single-view mask, we first apply inversion via iterative optimizations to find the corresponding latent vector for the preceding SOF~\cite{chen2019learning} network, which can generate novel view semantic masks for further image-to-image mapping. Note that SofGAN requires training data to have high-quality multi-view semantic masks, which is not available nor needed in our task. 
\textbf{pix2pixHD~\cite{wang2018high}} is an image translation approach. We adopt it with general GAN-inversion techniques~\cite{chan2021pi,karras2020analyzing}. For a given mask, it is first mapped to a photorealistic image via pix2pixHD, which will then be mapped to the corresponding latent code in $\pi$-GAN via GAN-inversion. With the recovered codes, multi-view images can be directly obtained using $\pi$-GAN.    
\textbf{pSp~\cite{richardson2021encoding}} is an image translation approach that is designed for encoding into StyleGAN2~\cite{karras2020analyzing}. We adapted it by using its ResNet~\cite{he2016deep}-based pSp encoder to replace the $\pi$-GAN mapping network, and we further trained the network with objective functions used by pSp.

\noindent\textbf{Evaluation Metrics.}
We show qualitative results by rendering images with different viewing directions and FOV (Field of View). 
We also report Frechet Inception Distance (FID)~\cite{heusel2017gans} and Inception Score (IS)~\cite{salimans2016improved} using Inception-v3~\cite{szegedy2016rethinking} over the test sets. 
Average running time and model sizes are also compared.

\noindent\textbf{Implementation Details.} Swin-T is used in all experiments with input resolution $224 \times 224$. For the decoder, the size of local region $\mathcal{R}$ is set to $128 \times 128$. The step size of each ray is set to 28. Other miscellaneous settings of the pre-trained decoder, \eg, ray depth ranges, are kept unchanged. Hyper-parameters in Eq.~\eqref{eq:total_loss} are set as $\lambda_{\text{rec}}$=1, $\lambda_{\text{LPIPS}}$=0.8, $\lambda_{\text{reg}}$=0.005, $\lambda_{\text{GAN}}$=0.08. The implementation is done in PyTorch~\cite{paszke2019pytorch}. More details are provided in Section~\ref{sec:tech_imple}.

\begin{figure}[t!]
    \centering
    \includegraphics[width=0.99\textwidth]{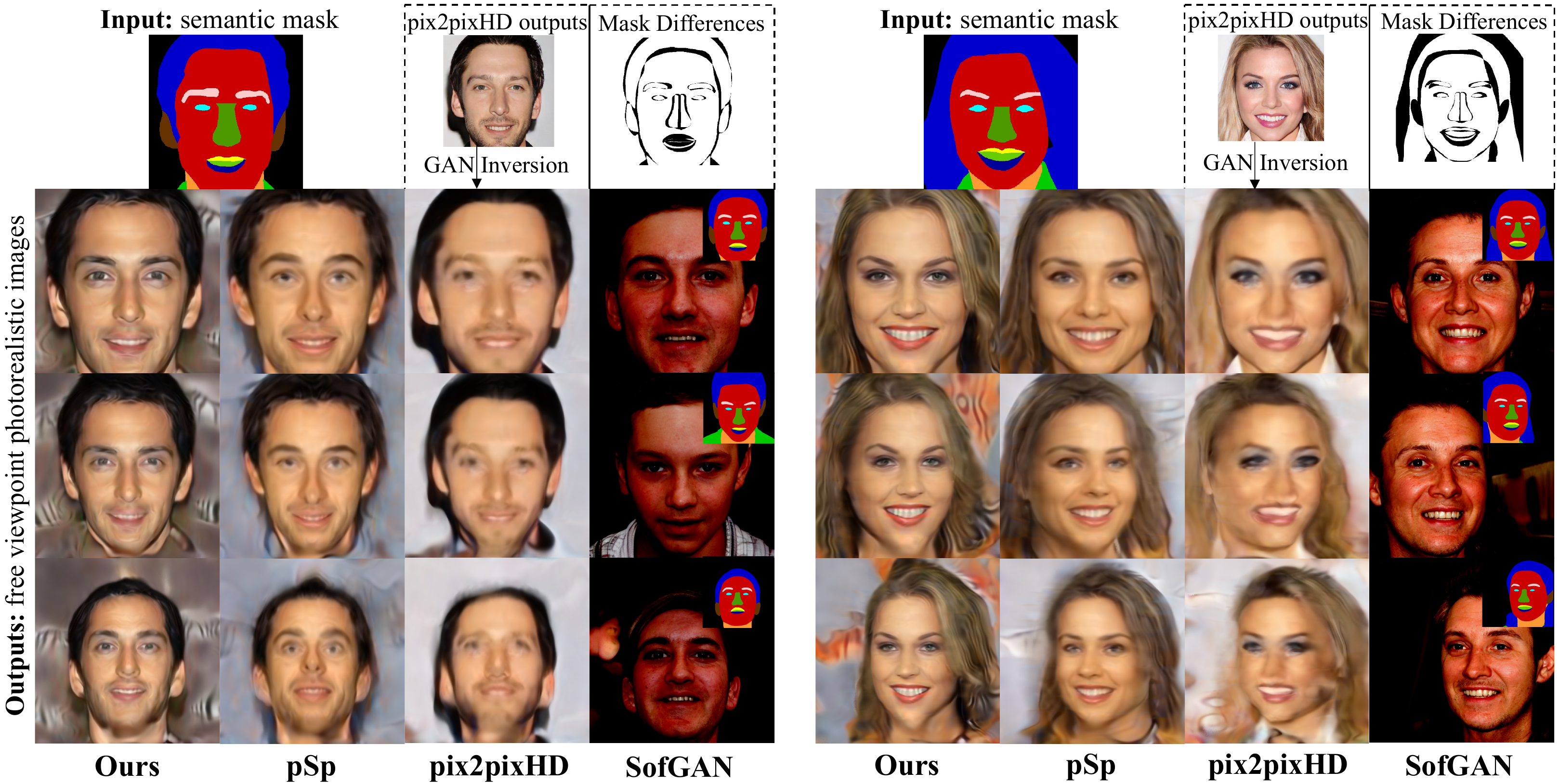} 
    \caption{Comparisons on CelebAMask. Images at each column are generated by the corresponding models mentioned at the bottom. Only SofGAN requires generation of multi-view semantic masks, shown at the top right corners of related images}
    \label{fig:compare_image}
\end{figure}

\subsection{Results}
\noindent\textbf{Comparisons on CelebAMask-HQ.} 
As shown in Fig.~\ref{fig:compare_image}, compared to all other baseline models,  \textbf{Sem2NeRF (1st\&5th columns)} achieved the best performance on both mapping accuracy and multi-view consistency. 
\textbf{pSp (2nd\&6th columns)} generated images with lower quality compared to ours, especially for novel views, mainly because our model is designed with a region-aware learning strategy and a GAN loss for random-posed images during training. The CNN-based encoder also failed to capture fine-grained details, \eg, eyebrow shapes for the left face. Our method and the inversion-based pix2pixHD were better in matching semantics compared to pSp.
\textbf{pix2pixHD (3rd\&7th columns)} can map semantic masks to high quality images in the same viewpoint (top row), but does not generate novel views well.
Basic GAN-inversion is not an efficient or easy way to find the desired latent codes, since the current 3D generative models are still quite immature. Even though images with the same viewing direction as the masks are reasonable, those novel view outputs contain artifacts.
\textbf{SofGAN (4th\&8th columns)} generated each single-view image with good quality; however, its results do not match with the given mask and lacked 3D consistency. The reason is that it is hard to map the given semantic mask to the desired latent codes of its semantic generator (SOF Net), whose sampling space is relatively small due to the lack of training data (only 122 subjects). The recovered mask did not match well with the given mask (top row). Besides, although the semantic masks show good multi-view consistency (top right corner of each image), conducting semantic-to-image mapping separately for each viewpoint does not guarantee that the consistency will be retained, since a semantic mask hardly contains any texture information and is geometrically ambiguous.

Quantitative results are give in Table~\ref{table:celeba}. 
It can be seen that our Sem2NeRF method achieves the best performance,  significantly outperforming the two baselines in both FID and IS scores. Note that we did not quantitatively compare single view image quality with SofGAN, considering that SofGAN for Semantic-to-NeRF is limited by its mask inversion quality and multi-view consistency, both of which cannot be measured by FID or IS scores.  We also notice that scores of all models are lower than expected. The main reason is that $\pi$-GAN is initially trained on CelebA, but due to the requirement of semantic masks, our task conducted experiments using CelebA-HQ. The domain gap between CelebA and CelebA-HQ reduced the FID scores dramatically. Besides, our model also sees advantages in terms of running time and model size. 

\begin{table}[t!]
\begin{center}
\caption{Quantitative comparisons on CelebAMask @ 128$\times$128} 
\label{table:celeba}
\begin{tabular*}{\textwidth}{l @{\extracolsep{\fill}} cccc}
\hline
 & FID~$\downarrow$  & IS~$\uparrow$ & Runtime(s)~$\downarrow$ & $\#$ Params(M)~$\downarrow$ \\
 \hline
pix2pixHD~\cite{wang2018high} (with inversion) & 67.32 & 1.72 &  161.59$\pm$0.859  & $\sim$184.24 \\
pSp~\cite{richardson2021encoding} & 55.56 & 1.74 & 0.25$\pm$0.004  & $\sim$138.27 \\
Sem2NeRF(Ours) & \textbf{41.52}  & \textbf{2.03}  & \textbf{0.18$\pm$0.003}  & \textbf{$\sim$32.01} \\
\hline
\end{tabular*} 
\end{center} 
\end{table}

\begin{figure}[t!]
    \centering
    \includegraphics[width=0.99\textwidth]{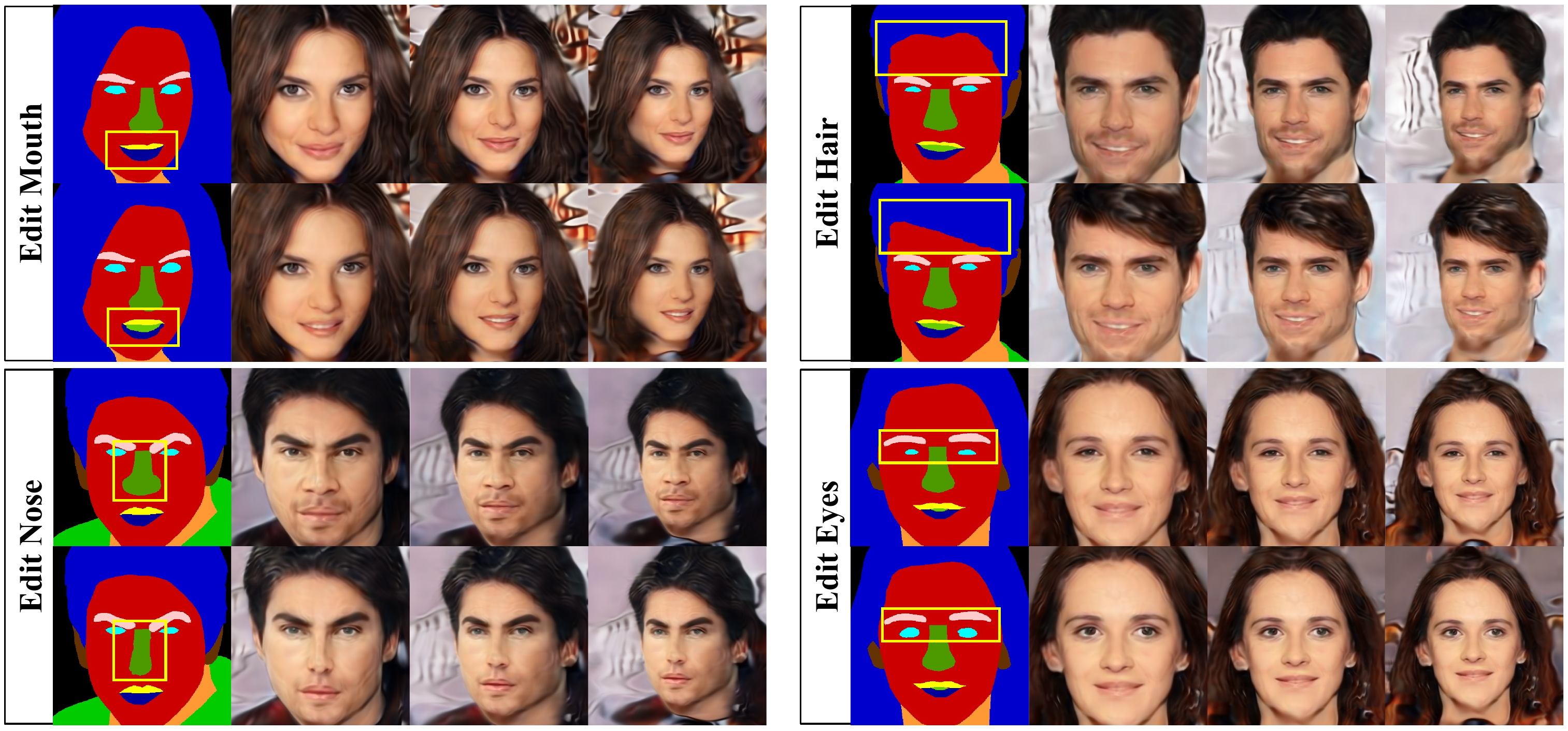}
    \caption{Editing 3D scenes by changing single-view semantic masks. Three viewpoints are shown for better comparison in each group}
    \label{fig:mask_edit}
\end{figure}

\noindent\textbf{Mask Editing.} As depicted in Fig.~\ref{fig:mask_edit}, our framework supports editing of 3D scenes by simply changing the given semantic mask, and is applicable to both labels associated with large regions, \eg, hair, as well as small regions, \eg, eyes, nose, mouth. 
This is not trivial since the semantic mask is not directly leveraged to control the 3D scene at the pixel level (if even possible), but is instead encoded into a sparse latent code, which may fail to preserve fine-grain editing. 
We address this challenge via the region-aware learning strategy. 

\begin{figure}[t!]
    \centering
    \includegraphics[width=0.99\textwidth]{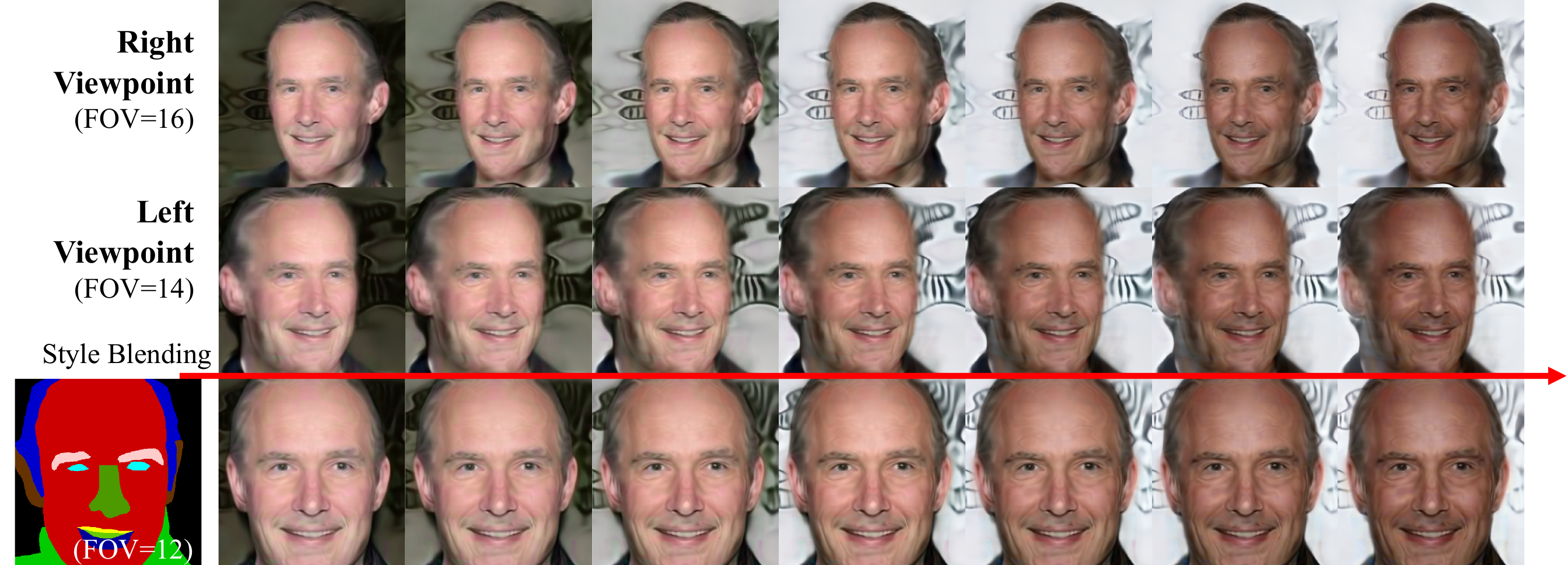}
    \caption{Multi-modal synthesis. Styles are linearly blended from left to right. Three viewpoints are provided from top to bottom}
    \label{fig:multimodal}
\end{figure}

\noindent\textbf{Multi-Modal Synthesis.} Sem2NeRF supports multi-modal synthesis by simply changing the last few layers of the style codes. As shown in Fig.~\ref{fig:multimodal}, we randomly sampled two style codes, and applied linear blending to continuously change the general styles of the 3D scenes generated by the given masks.

\noindent\textbf{Ablation Studies.}  
To further evaluate the efficacy of Sem2NeRF, we designed four ablation models, including 1) without region-aware encoder $\Phi_{\text{wo\_RE}}$, where the Swin-T encoder is replaced by the pSp encoder; 2) without region-aware decoder $\Phi_{\text{wo\_RD}}$, where the region-aware ray sampling strategy is discarded; 3) without input augmentation $\Phi_{\text{wo\_IA}}$, where contours and distance field representations are removed from the input; and 4) without random-pose GAN loss $\Phi_{\text{wo\_GAN}}$, where both Eq.~\eqref{eq:gan} and the discriminator are removed.

\begin{figure}[t!]
    \centering
    \includegraphics[width=0.99\textwidth]{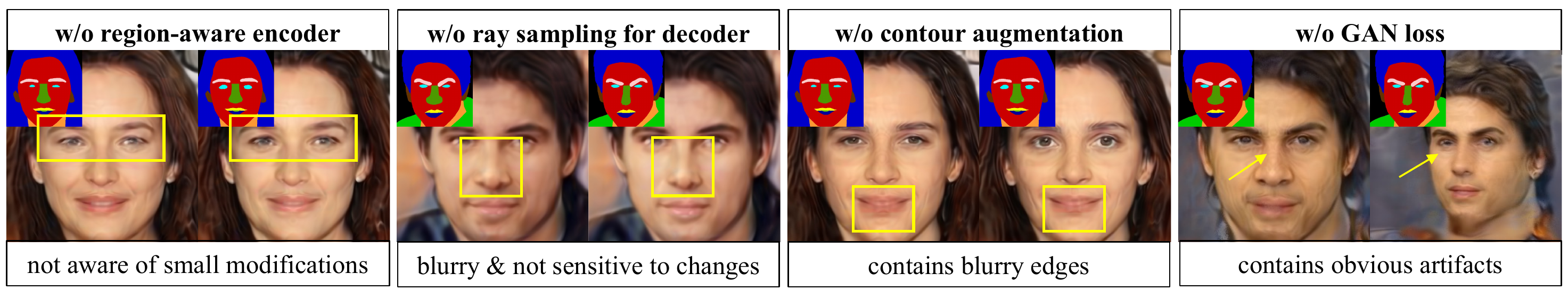}
    \caption{Results of ablation studies. Each group (two views) is generated by a model without the component mentioned at the top. Main issues are described at the bottom}
    \label{fig:ablation}
\end{figure}

As shown in Fig.~\ref{fig:ablation}, compared to the full model (Fig.~\ref{fig:mask_edit}), $\Phi_{\text{wo\_RE}}$ (1st group) is not sensitive to changes in small regions, \ie, eyes, mainly because the CNN-based encoder tends to ignore small changes. $\Phi_{\text{wo\_RD}}$ (2nd group) shows similar pattern (nose region) as the latent codes are not trained to be region-aware. It also has lower image quality, because the region-aware strategy enables denser sampling. $\Phi_{\text{wo\_IA}}$ (3rd group) achieves comparable performance but with blurry edges for some regions, \eg, mouth. This is because both contour and distance field representation help highlight the boundary information. Finally, images obtained by $\Phi_{\text{wo\_GAN}}$ (4th group) have more artifacts in both views, demonstrating that GAN loss is important for improving the image quality of different poses.

\begin{figure}[t!]
    \centering
    \includegraphics[width=0.99\textwidth]{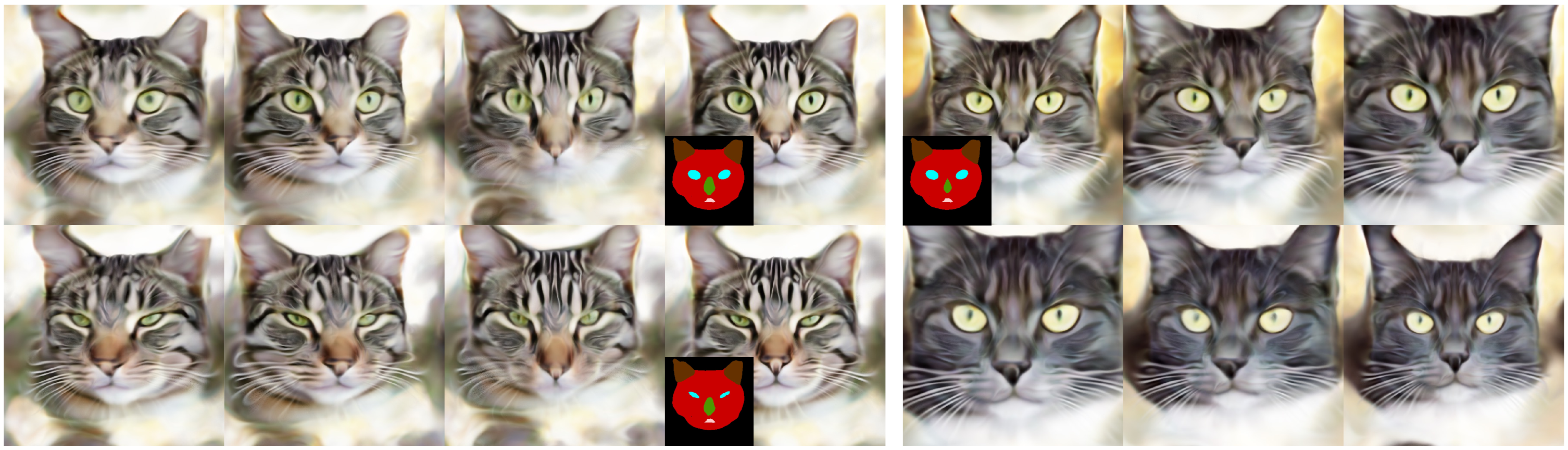}
    \caption{Results on CatMask. Left part compares results of changing eyes shape. Right part showcases results of style linear blending (in zigzag order)}
    \label{fig:cat_cases}
\end{figure}

\noindent\textbf{Experiments beyond Human Faces.} The introduced task can easily go beyond the human face domain by leveraging state-of-the-art weakly supervised semantic segmentation model to create pseudo labels. In this work, we present a Cat face example.
Experimental results are shown in Fig.~\ref{fig:cat_cases}. 
Even when training with noisy pseudo labels, Sem2NeRF is robust enough to generate plausible results. For a given cat semantic mask, our model can map it to a 3D scene and render cat faces from arbitrary viewpoints, including different viewing directions (left part), and different FOV (right part). It also allows changing the 3D scenes by editing the single-view semantic masks, \eg, changing the eye shape (left two rows). 
Multi-modal synthesis is also supported (right part in zigzag order).

\begin{figure}[t!]
    \centering
    \includegraphics[width=0.99\textwidth]{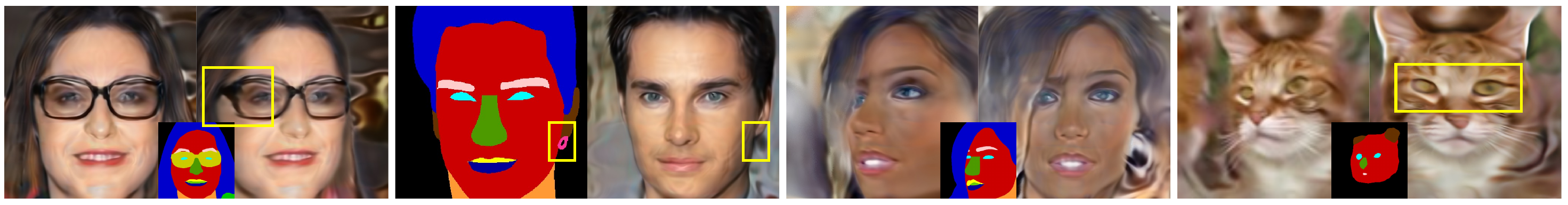}
    \caption{Challenging cases of Sem2NeRF on Semantic-to-NeRF translation}
    \label{fig:fail_cases}
\end{figure}

\noindent\textbf{Challenging Cases.} Although Sem2NeRF addresses the Semantic-to-NeRF task in most cases, its advantages rely on an assumption, namely the generative capability of the pre-trained decoder. We show some challenging cases in Fig.~\ref{fig:fail_cases}. Accessories may have the wrong geometric shape (glasses in 1st case), or fail to render (earring in 2nd case), while masks with extreme poses might be converted to 3D scenes with abnormal texture or distorted contents (last two cases).

\section{Conclusions}

We have presented an initial step of extending the 2D image-to-image task to the 3D space, and introduced a new task called Semantic-to-NeRF translation. It aims to reconstruct a NeRF-based 3D scene, by taking as input only one single-view semantic mask. We further proposed Sem2NeRF model, which addresses the task via encoding the semantic mask into the latent space of a pre-trained 3D generative model.
More importantly, we intriguingly found the importance of regional awareness for this new task, and  tamed Sem2NeRF with a region-aware learning strategy. We demonstrated the capability of Sem2NeRF regarding free viewpoint generation, mask editing and multi-modal synthesis on two benchmark datasets, and showcased the superiority of our framework compared to three strong baselines. Future work will include adding more scenarios to the new task, and supporting changing styles for specific regions.

\noindent \textbf{Acknowledgements} This research is partially supported by CRC Building 4.0 Project \#44.

\clearpage
%
%
\bibliographystyle{splncs04}
\bibliography{egbib}

\begin{thebibliography}{10}
\providecommand{\url}[1]{\texttt{#1}}
\providecommand{\urlprefix}{URL }
\providecommand{\doi}[1]{https://doi.org/#1}

\bibitem{barron2021mip}
Barron, J.T., Mildenhall, B., Tancik, M., Hedman, P., Martin-Brualla, R.,
  Srinivasan, P.P.: Mip-nerf: A multiscale representation for anti-aliasing
  neural radiance fields. In: Int. Conf. Comput. Vis. pp. 5855--5864 (2021)

\bibitem{chan2021efficient}
Chan, E.R., Lin, C.Z., Chan, M.A., Nagano, K., Pan, B., De~Mello, S., Gallo,
  O., Guibas, L., Tremblay, J., Khamis, S., et~al.: Efficient geometry-aware 3d
  generative adversarial networks. IEEE Conf. Comput. Vis. Pattern Recog.
  (2022)

\bibitem{chan2021pi}
Chan, E.R., Monteiro, M., Kellnhofer, P., Wu, J., Wetzstein, G.: pi-gan:
  Periodic implicit generative adversarial networks for 3d-aware image
  synthesis. In: IEEE Conf. Comput. Vis. Pattern Recog. pp. 5799--5809 (2021)

\bibitem{chen2022sofgan}
Chen, A., Liu, R., Xie, L., Chen, Z., Su, H., Yu, J.: Sofgan: A portrait image
  generator with dynamic styling. ACM Trans. Graph.  \textbf{41}(1),  1--26
  (2022)

\bibitem{chen2017deeplab}
Chen, L.C., Papandreou, G., Kokkinos, I., Murphy, K., Yuille, A.L.: Deeplab:
  Semantic image segmentation with deep convolutional nets, atrous convolution,
  and fully connected crfs. IEEE Trans. Pattern Anal. Mach. Intell.
  \textbf{40}(4),  834--848 (2017)

\bibitem{chen2018sketchygan}
Chen, W., Hays, J.: Sketchygan: Towards diverse and realistic sketch to image
  synthesis. In: IEEE Conf. Comput. Vis. Pattern Recog. pp. 9416--9425 (2018)

\bibitem{chen2021edge}
Chen, Y., Huang, J., Wang, J., Xie, X.: Edge prior augmented networks for
  motion deblurring on naturally blurry images. arXiv preprint arXiv:2109.08915
   (2021)

\bibitem{chen2019learning}
Chen, Z., Zhang, H.: Learning implicit fields for generative shape modeling.
  In: IEEE Conf. Comput. Vis. Pattern Recog. pp. 5939--5948 (2019)

\bibitem{choi2020stargan}
Choi, Y., Uh, Y., Yoo, J., Ha, J.W.: Stargan v2: Diverse image synthesis for
  multiple domains. In: IEEE Conf. Comput. Vis. Pattern Recog. pp. 8188--8197
  (2020)

\bibitem{collins2020editing}
Collins, E., Bala, R., Price, B., Susstrunk, S.: Editing in style: Uncovering
  the local semantics of gans. In: IEEE Conf. Comput. Vis. Pattern Recog. pp.
  5771--5780 (2020)

\bibitem{deng2009imagenet}
Deng, J., Dong, W., Socher, R., Li, L.J., Li, K., Fei-Fei, L.: Imagenet: A
  large-scale hierarchical image database. In: IEEE Conf. Comput. Vis. Pattern
  Recog. pp. 248--255. Ieee (2009)

\bibitem{dosovitskiy2020image}
Dosovitskiy, A., Beyer, L., Kolesnikov, A., Weissenborn, D., Zhai, X.,
  Unterthiner, T., Dehghani, M., Minderer, M., Heigold, G., Gelly, S., et~al.:
  An image is worth 16x16 words: Transformers for image recognition at scale.
  Int. Conf. Learn. Represent.  (2021)

\bibitem{goel2020shape}
Goel, S., Kanazawa, A., Malik, J.: Shape and viewpoint without keypoints. In:
  Eur. Conf. Comput. Vis. pp. 88--104. Springer (2020)

\bibitem{goodfellow2014generative}
Goodfellow, I., Pouget-Abadie, J., Mirza, M., Xu, B., Warde-Farley, D., Ozair,
  S., Courville, A., Bengio, Y.: Generative adversarial nets. Adv. Neural
  Inform. Process. Syst.  \textbf{27} (2014)

\bibitem{gu2021stylenerf}
Gu, J., Liu, L., Wang, P., Theobalt, C.: Stylenerf: A style-based 3d-aware
  generator for high-resolution image synthesis. Int. Conf. Learn. Represent.
  (2022)

\bibitem{hao2021gancraft}
Hao, Z., Mallya, A., Belongie, S., Liu, M.Y.: Gancraft: Unsupervised 3d neural
  rendering of minecraft worlds. In: Int. Conf. Comput. Vis. pp. 14072--14082
  (2021)

\bibitem{he2016deep}
He, K., Zhang, X., Ren, S., Sun, J.: Deep residual learning for image
  recognition. In: IEEE Conf. Comput. Vis. Pattern Recog. pp. 770--778 (2016)

\bibitem{heusel2017gans}
Heusel, M., Ramsauer, H., Unterthiner, T., Nessler, B., Hochreiter, S.: Gans
  trained by a two time-scale update rule converge to a local nash equilibrium.
  Adv. Neural Inform. Process. Syst.  \textbf{30} (2017)

\bibitem{huang2020semantic}
Huang, H.P., Tseng, H.Y., Lee, H.Y., Huang, J.B.: Semantic view synthesis. In:
  Eur. Conf. Comput. Vis. pp. 592--608. Springer (2020)

\bibitem{isola2017image}
Isola, P., Zhu, J.Y., Zhou, T., Efros, A.A.: Image-to-image translation with
  conditional adversarial networks. In: IEEE Conf. Comput. Vis. Pattern Recog.
  pp. 1125--1134 (2017)

\bibitem{johnson2016perceptual}
Johnson, J., Alahi, A., Fei-Fei, L.: Perceptual losses for real-time style
  transfer and super-resolution. In: Eur. Conf. Comput. Vis. pp. 694--711.
  Springer (2016)

\bibitem{kanazawa2018learning}
Kanazawa, A., Tulsiani, S., Efros, A.A., Malik, J.: Learning category-specific
  mesh reconstruction from image collections. In: Eur. Conf. Comput. Vis. pp.
  371--386 (2018)

\bibitem{karras2017progressive}
Karras, T., Aila, T., Laine, S., Lehtinen, J.: Progressive growing of gans for
  improved quality, stability, and variation. Int. Conf. Learn. Represent.
  (2018)

\bibitem{karras2019style}
Karras, T., Laine, S., Aila, T.: A style-based generator architecture for
  generative adversarial networks. In: IEEE Conf. Comput. Vis. Pattern Recog.
  pp. 4401--4410 (2019)

\bibitem{karras2020analyzing}
Karras, T., Laine, S., Aittala, M., Hellsten, J., Lehtinen, J., Aila, T.:
  Analyzing and improving the image quality of stylegan. In: IEEE Conf. Comput.
  Vis. Pattern Recog. pp. 8110--8119 (2020)

\bibitem{kingma2014adam}
Kingma, D.P., Ba, J.: Adam: A method for stochastic optimization. In: Int.
  Conf. Learn. Represent. (2014)

\bibitem{lee2020maskgan}
Lee, C.H., Liu, Z., Wu, L., Luo, P.: Maskgan: Towards diverse and interactive
  facial image manipulation. In: IEEE Conf. Comput. Vis. Pattern Recog. pp.
  5549--5558 (2020)

\bibitem{levoy1990efficient}
Levoy, M.: Efficient ray tracing of volume data. ACM Trans. Graph.
  \textbf{9}(3),  245--261 (1990)

\bibitem{ling2021editgan}
Ling, H., Kreis, K., Li, D., Kim, S.W., Torralba, A., Fidler, S.: Editgan:
  High-precision semantic image editing. In: Adv. Neural Inform. Process. Syst.
  (2021)

\bibitem{lira2020ganhopper}
Lira, W., Merz, J., Ritchie, D., Cohen-Or, D., Zhang, H.: Ganhopper: Multi-hop
  gan for unsupervised image-to-image translation. In: Eur. Conf. Comput. Vis.
  pp. 363--379. Springer (2020)

\bibitem{liu2020neural}
Liu, L., Gu, J., Zaw~Lin, K., Chua, T.S., Theobalt, C.: Neural sparse voxel
  fields. Adv. Neural Inform. Process. Syst.  \textbf{33},  15651--15663 (2020)

\bibitem{liu2022semantic}
Liu, X., Xu, Y., Wu, Q., Zhou, H., Wu, W., Zhou, B.: Semantic-aware implicit
  neural audio-driven video portrait generation. Eur. Conf. Comput. Vis.
  (2022)

\bibitem{liu2021swin}
Liu, Z., Lin, Y., Cao, Y., Hu, H., Wei, Y., Zhang, Z., Lin, S., Guo, B.: Swin
  transformer: Hierarchical vision transformer using shifted windows. In: Int.
  Conf. Comput. Vis. pp. 10012--10022 (2021)

\bibitem{liu2015deep}
Liu, Z., Luo, P., Wang, X., Tang, X.: Deep learning face attributes in the
  wild. In: Int. Conf. Comput. Vis. pp. 3730--3738 (2015)

\bibitem{lombardi2021mixture}
Lombardi, S., Simon, T., Schwartz, G., Zollhoefer, M., Sheikh, Y., Saragih, J.:
  Mixture of volumetric primitives for efficient neural rendering. ACM Trans.
  Graph.  \textbf{40}(4),  1--13 (2021)

\bibitem{luo2021diffusion}
Luo, S., Hu, W.: Diffusion probabilistic models for 3d point cloud generation.
  In: IEEE Conf. Comput. Vis. Pattern Recog. pp. 2837--2845 (2021)

\bibitem{mescheder2018training}
Mescheder, L., Geiger, A., Nowozin, S.: Which training methods for gans do
  actually converge? In: Int. Conf. Mach. Learn. pp. 3481--3490. PMLR (2018)

\bibitem{mildenhall2020nerf}
Mildenhall, B., Srinivasan, P.P., Tancik, M., Barron, J.T., Ramamoorthi, R.,
  Ng, R.: Nerf: Representing scenes as neural radiance fields for view
  synthesis. In: Eur. Conf. Comput. Vis. pp. 405--421. Springer (2020)

\bibitem{niemeyer2021giraffe}
Niemeyer, M., Geiger, A.: Giraffe: Representing scenes as compositional
  generative neural feature fields. In: IEEE Conf. Comput. Vis. Pattern Recog.
  pp. 11453--11464 (2021)

\bibitem{park2019semantic}
Park, T., Liu, M.Y., Wang, T.C., Zhu, J.Y.: Semantic image synthesis with
  spatially-adaptive normalization. In: IEEE Conf. Comput. Vis. Pattern Recog.
  pp. 2337--2346 (2019)

\bibitem{paszke2019pytorch}
Paszke, A., Gross, S., Massa, F., Lerer, A., Bradbury, J., Chanan, G., Killeen,
  T., Lin, Z., Gimelshein, N., Antiga, L., et~al.: Pytorch: An imperative
  style, high-performance deep learning library. Adv. Neural Inform. Process.
  Syst.  \textbf{32} (2019)

\bibitem{perez2018film}
Perez, E., Strub, F., De~Vries, H., Dumoulin, V., Courville, A.: Film: Visual
  reasoning with a general conditioning layer. In: AAAI. vol.~32 (2018)

\bibitem{pumarola2021d}
Pumarola, A., Corona, E., Pons-Moll, G., Moreno-Noguer, F.: D-nerf: Neural
  radiance fields for dynamic scenes. In: IEEE Conf. Comput. Vis. Pattern
  Recog. pp. 10318--10327 (2021)

\bibitem{richardson2021encoding}
Richardson, E., Alaluf, Y., Patashnik, O., Nitzan, Y., Azar, Y., Shapiro, S.,
  Cohen-Or, D.: Encoding in style: a stylegan encoder for image-to-image
  translation. In: IEEE Conf. Comput. Vis. Pattern Recog. pp. 2287--2296 (2021)

\bibitem{salimans2016improved}
Salimans, T., Goodfellow, I., Zaremba, W., Cheung, V., Radford, A., Chen, X.:
  Improved techniques for training gans. Adv. Neural Inform. Process. Syst.
  \textbf{29} (2016)

\bibitem{schwarz2020graf}
Schwarz, K., Liao, Y., Niemeyer, M., Geiger, A.: Graf: Generative radiance
  fields for 3d-aware image synthesis. Adv. Neural Inform. Process. Syst.
  \textbf{33},  20154--20166 (2020)

\bibitem{shi2022semanticstylegan}
Shi, Y., Yang, X., Wan, Y., Shen, X.: Semanticstylegan: Learning compositional
  generative priors for controllable image synthesis and editing. In: IEEE
  Conf. Comput. Vis. Pattern Recog. pp. 11254--11264 (2022)

\bibitem{sitzmann2020implicit}
Sitzmann, V., Martel, J., Bergman, A., Lindell, D., Wetzstein, G.: Implicit
  neural representations with periodic activation functions. Adv. Neural
  Inform. Process. Syst.  \textbf{33},  7462--7473 (2020)

\bibitem{song2021agilegan}
Song, G., Luo, L., Liu, J., Ma, W.C., Lai, C., Zheng, C., Cham, T.J.: Agilegan:
  stylizing portraits by inversion-consistent transfer learning. ACM Trans.
  Graph.  \textbf{40}(4),  1--13 (2021)

\bibitem{sun2021direct}
Sun, C., Sun, M., Chen, H.T.: Direct voxel grid optimization: Super-fast
  convergence for radiance fields reconstruction. IEEE Conf. Comput. Vis.
  Pattern Recog.  (2022)

\bibitem{sun2022fenerf}
Sun, J., Wang, X., Zhang, Y., Li, X., Zhang, Q., Liu, Y., Wang, J.: Fenerf:
  Face editing in neural radiance fields. In: IEEE Conf. Comput. Vis. Pattern
  Recog. pp. 7672--7682 (2022)

\bibitem{szegedy2016rethinking}
Szegedy, C., Vanhoucke, V., Ioffe, S., Shlens, J., Wojna, Z.: Rethinking the
  inception architecture for computer vision. In: IEEE Conf. Comput. Vis.
  Pattern Recog. pp. 2818--2826 (2016)

\bibitem{wang2018high}
Wang, T.C., Liu, M.Y., Zhu, J.Y., Tao, A., Kautz, J., Catanzaro, B.:
  High-resolution image synthesis and semantic manipulation with conditional
  gans. In: IEEE Conf. Comput. Vis. Pattern Recog. pp. 8798--8807 (2018)

\bibitem{wu2016learning}
Wu, J., Zhang, C., Xue, T., Freeman, B., Tenenbaum, J.: Learning a
  probabilistic latent space of object shapes via 3d generative-adversarial
  modeling. Adv. Neural Inform. Process. Syst.  \textbf{29} (2016)

\bibitem{wu2022objectsdf}
Wu, Q., Liu, X., Chen, Y., Li, K., Zheng, C., Cai, J., Zheng, J.:
  Object-compositional neural implicit surfaces. Eur. Conf. Comput. Vis.
  (2022)

\bibitem{wu2021stylespace}
Wu, Z., Lischinski, D., Shechtman, E.: Stylespace analysis: Disentangled
  controls for stylegan image generation. In: IEEE Conf. Comput. Vis. Pattern
  Recog. pp. 12863--12872 (2021)

\bibitem{wu2021stylealign}
Wu, Z., Nitzan, Y., Shechtman, E., Lischinski, D.: Stylealign: Analysis and
  applications of aligned stylegan models. Int. Conf. Learn. Represent.  (2022)

\bibitem{xu2022transeditor}
Xu, Y., Yin, Y., Jiang, L., Wu, Q., Zheng, C., Loy, C.C., Dai, B., Wu, W.:
  Transeditor: Transformer-based dual-space gan for highly controllable facial
  editing. In: IEEE Conf. Comput. Vis. Pattern Recog. pp. 7683--7692 (2022)

\bibitem{xu20213d}
Xu, Y., Peng, S., Yang, C., Shen, Y., Zhou, B.: 3d-aware image synthesis via
  learning structural and textural representations. IEEE Conf. Comput. Vis.
  Pattern Recog.  (2022)

\bibitem{yen2021inerf}
Yen-Chen, L., Florence, P., Barron, J.T., Rodriguez, A., Isola, P., Lin, T.Y.:
  inerf: Inverting neural radiance fields for pose estimation. In: IEEE Int.
  Conf. Intell. Robots Syst. pp. 1323--1330. IEEE (2021)

\bibitem{zhang20213d}
Zhang, J., Sangineto, E., Tang, H., Siarohin, A., Zhong, Z., Sebe, N., Wang,
  W.: 3d-aware semantic-guided generative model for human synthesis. Eur. Conf.
  Comput. Vis.  (2022)

\bibitem{zhang2019making}
Zhang, R.: Making convolutional networks shift-invariant again. In: Int. Conf.
  Mach. Learn. pp. 7324--7334. PMLR (2019)

\bibitem{zhang2018unreasonable}
Zhang, R., Isola, P., Efros, A.A., Shechtman, E., Wang, O.: The unreasonable
  effectiveness of deep features as a perceptual metric. In: IEEE Conf. Comput.
  Vis. Pattern Recog. pp. 586--595 (2018)

\bibitem{zhang2022multi}
Zhang, X., Zheng, Z., Gao, D., Zhang, B., Pan, P., Yang, Y.: Multi-view
  consistent generative adversarial networks for 3d-aware image synthesis. In:
  IEEE Conf. Comput. Vis. Pattern Recog. pp. 18450--18459 (2022)

\bibitem{zhang2021datasetgan}
Zhang, Y., Ling, H., Gao, J., Yin, K., Lafleche, J.F., Barriuso, A., Torralba,
  A., Fidler, S.: Datasetgan: Efficient labeled data factory with minimal human
  effort. In: IEEE Conf. Comput. Vis. Pattern Recog. pp. 10145--10155 (2021)

\bibitem{zheng2019pluralistic}
Zheng, C., Cham, T.J., Cai, J.: Pluralistic image completion. In: IEEE Conf.
  Comput. Vis. Pattern Recog. pp. 1438--1447 (2019)

\bibitem{zheng2021tfill}
Zheng, C., Cham, T.J., Cai, J.: Tfill: Image completion via a transformer-based
  architecture. IEEE Conf. Comput. Vis. Pattern Recog.  (2022)

\bibitem{zhi2021place}
Zhi, S., Laidlow, T., Leutenegger, S., Davison, A.J.: In-place scene labelling
  and understanding with implicit scene representation. In: Int. Conf. Comput.
  Vis. pp. 15838--15847 (2021)

\bibitem{zhou2018stereo}
Zhou, T., Tucker, R., Flynn, J., Fyffe, G., Snavely, N.: Stereo magnification:
  Learning view synthesis using multiplane images. ACM Trans. Graph.  (2018)

\bibitem{zhu2017unpaired}
Zhu, J.Y., Park, T., Isola, P., Efros, A.A.: Unpaired image-to-image
  translation using cycle-consistent adversarial networks. In: Int. Conf.
  Comput. Vis. pp. 2223--2232 (2017)

\bibitem{zhu2017toward}
Zhu, J.Y., Zhang, R., Pathak, D., Darrell, T., Efros, A.A., Wang, O.,
  Shechtman, E.: Toward multimodal image-to-image translation. Adv. Neural
  Inform. Process. Syst.  \textbf{30} (2017)

\end{thebibliography}

\clearpage

\appendix
\section{Additional Technical Details} \label{sec:tech}

\subsection{Builders for Additional Inputs} \label{sec:tech_input}

As mentioned in Section~\ref{sec:method_enc}, additional inputs, \ie, contour and distance field representation, are provided for the encoder of Sem2NeRF to further highlight the boundary information. In this subsection, we will detail how to build these data using the python package ``cv2''\footnote{\url{https://pypi.org/project/opencv-python/}}. Note that both of them are directly calculated from the semantic mask, without involving any extra labels. And each builder function can be done in $1\sim2$ milliseconds.

\begin{figure}[t!]
    \centering
    \includegraphics[width=0.99\textwidth]{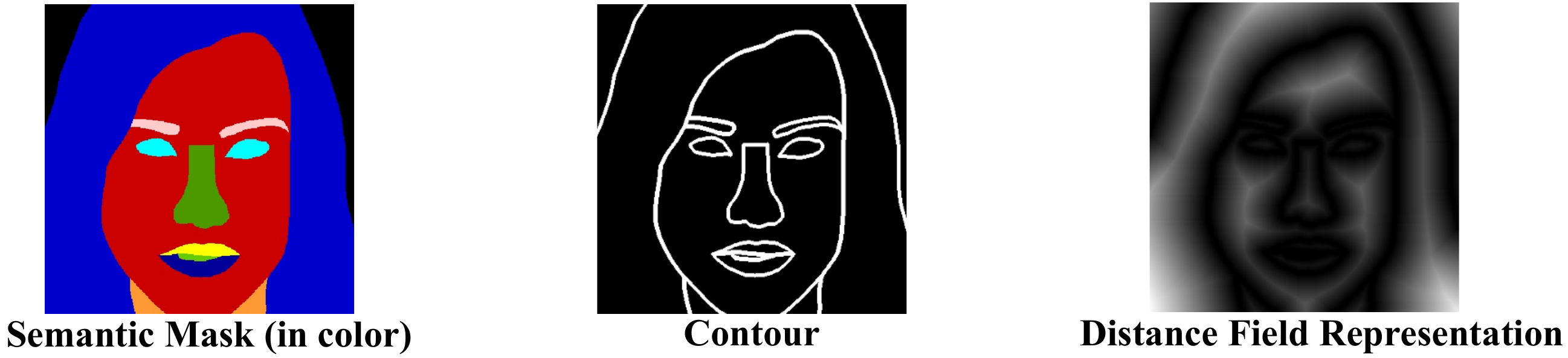}
    \caption{An example of the encoder input. Semantic mask is shown in color for better visualization, while the network takes one-hot encoded mask as input}
    \label{fig:input_aug}
\end{figure}

\textbf{Contour} is generally represented as a curve, joining all the continuous points that share the same intensity. It is widely used as a tool to help shape analysis, object detection, \etc. In our implementation, we build the contour from the given one-hot encoded semantic mask. Main python codes are given as below. An output example is shown in Figure~\ref{fig:input_aug} (middle).

\begin{lstlisting}[language=Python]
# --------------------- CONTOUR BUILDER ---------------------
def binary_masks_to_contour(binary_masks):
    ''' INPUT: semantic mask in one-hot encoding form
        OUTPUT: a contour map of the given semantic mask '''

    # initialize a black canvas
    mask = numpy.zeros((512, 512, 3), dtype=numpy.uint8)
    # find contours for each label 
    for binary_mask in binary_masks:
        cnts = cv2.findContours(binary_mask, 
                                cv2.RETR_EXTERNAL, 
                                cv2.CHAIN_APPROX_SIMPLE)
        cnts = cnts[0] if len(cnts) == 2 else cnts[1]
        for c in cnts:
            # draw contour with white color on the canvas
            cv2.drawContours(mask, [c], -1, (255, 255, 255), 
                             thickness=3)
    contour = cv2.cvtColor(mask, cv2.COLOR_BGR2GRAY)
    return contour
# --------------------------- END ---------------------------
\end{lstlisting}

\textbf{Distance field representation} is a dense representation extracted from binary image via distance transformation. In the distance field, the grey intensity of each pixel indicates its distance to the nearest boundary. An unsigned Euclidean distance field representation is adopted in our experiments. Main python codes are given as below. An output example is shown in Figure~\ref{fig:input_aug} (right).

\begin{lstlisting}[language=Python]
# ---------- DISTANCE FIELD REPRESENTATION BUILDER ----------
def contour_to_dist_filed(contour):
    ''' INPUT: contour, contour of the semantic mask
        OUTPUT: dist_field, distance filed representation '''

    # invert background and foreground of contour to match the setting
    invert_contour = cv2.bitwise_not(contour)
    dist_field = cv2.distanceTransform(invert_contour, 
                                       cv2.DIST_L2, 3)
    # normalize the distance filed to [0, 1]
    cv2.normalize(dist_field, dist_field, 0, 1.0, 
                  cv2.NORM_MINMAX)
    dist_field = dist_field * 255.
    return dist_field
# --------------------------- END --------------------------
\end{lstlisting}

\subsection{CatMask Dataset Rendering} \label{sec:tech_cat}

To better demonstrate the introduced Semantic-to-NeRF translation task and evaluate the proposed Sem2NeRF framework, we develop a general solution to create datasets with pseudo labels using minimal human effort. In this work, we present an example by rendering a cat faces dataset, termed CatMask, which contains single-view cat faces generated by the pre-trained $\pi$-GAN model, pseudo ground-truth viewing directions, and 6-classes semantic masks labelled by DatasetGAN~\cite{zhang2021datasetgan}. 
Similar to CelebAMask-HQ, CatMask only varies on the yaw and pitch axis. And it contains 28,000 training images and 2,000 testing images.

\textbf{Cat faces from $\pi$-GAN.} As mentioned in Section~\ref{sec:exp_set}, we assume the training data to have the viewing directions / poses of the single-view semantic-image pairs. However, different from human face, it is difficult to get the poses for cat faces. Considering that a pretrained $\pi$-GAN can generate photorealistic cat faces with given random poses, we choose to generate pseudo data for training our Sem2NeRF. Specifically, we randomly sample 30,000 vectors $\mathbf{z}\in\mathbb{R}^{256}$ from the input distribution of $\pi$-GAN to generate 30,000 corresponding cat faces by using the released pretrained model\footnote{\url{https://github.com/marcoamonteiro/pi-GAN}}. Each image comes with one  viewing direction, randomly sampled from normal distributions, with $X \sim \mathcal{N}(\pi /2,\,0.3^{2})$ for the yaw axis, $X \sim \mathcal{N}(\pi /2,\,0.1^{2})$ for the pitch axis, and 0 for the roll axis. Camera FOV is set to 18 to ensure the generated image covering the full cat face. Ray sampling resolution is set to $512\times512$, with ray depth range [0.8, 1.2] and ray step size 72. Hierarchical sampling is enabled to improve image quality. We save the rendering viewing direction and the generated images for the CatMask dataset.

\textbf{Cat semantic by DatasetGAN.} A suitable training dataset for Sem2NeRF should be able to be modelled by existing NeRF-based generator, while also comes with semantic labels. 
However, most datasets used by existing 3D-aware generative models, \eg, cat and car, do not contain component-level semantic masks. We further find out that DatasetGAN can create reasonable semantic masks labels for our task.

DatasetGAN is introduced to automatically build datasets of high-quality semantically segmented images.
Specifically, it proposes a MLP-based ``Style Interpreter'' that can be trained to decode the feature maps of a pretrained StyleGAN model to semantic labels, requiring only very few manually-labeled training samples, \eg, 30 labelled images for cat dataset. In this case, dataset can be automatically built by first randomly sampling images from StyleGAN, following by parsing with the trained style interpreter to obtain corresponding semantic labels. 

\begin{figure}[t!]
    \centering
    \includegraphics[width=0.99\textwidth]{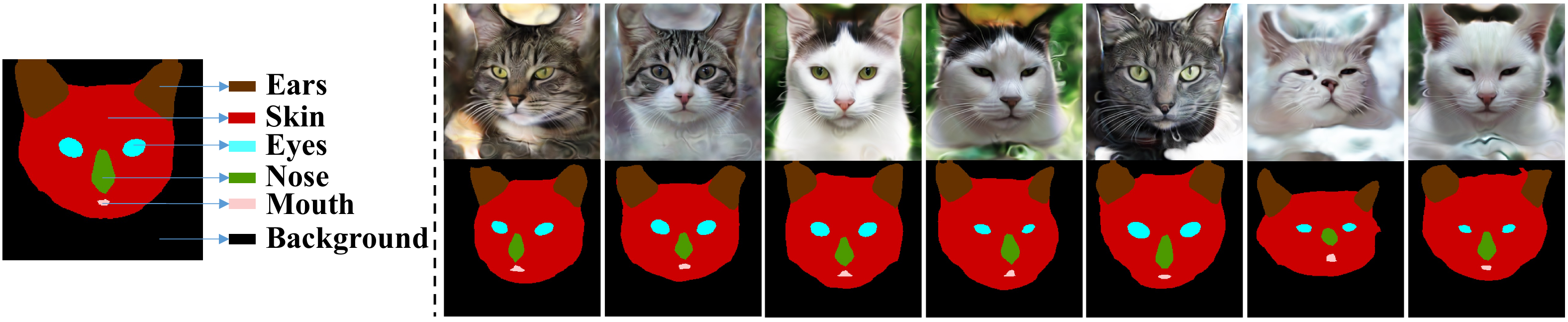}
    \caption{CatMask dataset. Left: label legends of 6 semantic classes. Right: single-view cat faces rendered by $\pi$-GAN (top row) and the corresponding semantic masks labelled by DatasetGAN (bottom row). Best view in high quality color image}
    \label{fig:catmask_dataset}
\end{figure}

We use the released\footnote{\url{https://github.com/nv-tlabs/datasetGAN\_release}} pretrained style interpreter for cat to generate a dataset of 10,000 images-semantic pairs. Such a dataset is further leveraged to train a Deeplab-V3~\cite{chen2017deeplab} model, which takes a cat image as input and outputs the corresponding cat semantic mask. We then use the trained Deeplab-V3 to label our generated CatMask dataset. 6 classes are selected based on the label quality. Label legends and examples of the CatMask dataset are given in Figure~\ref{fig:catmask_dataset}.

\subsection{Additional Implementation Details} \label{sec:tech_imple}

\textbf{Style codes averages ($\overline{\gamma}, \overline{\beta}$).} We randomly sample 10,000 vectors $\mathbf{z}\in\mathbb{R}^{256}$ from a standard normal distribution, then feed them through the pretrained mapping network of the origianl $\pi$-GAN model, finally average the outputs over the batch dimension to obtain $\overline{\gamma}, \overline{\beta}$.

\textbf{Datasets.} For CelebAMask-HQ~\cite{lee2020maskgan}, both images and semantic masks are loaded as resolution $640\times640$, then center cropped to $512\times512$. For CatMask, images and semantic masks are directly loaded as $512\times512$. Semantic masks are transformed using one-hot encoding, augmented with the aforementioned contours and distance field representations. We do not apply any other data augmentation, \eg, random flip, to avoid harming the pose information.

\textbf{Training.} Patch discriminator with input size $128\times128$ is adopted in our experiments, using the implementation provided by the GRAF~\cite{schwarz2020graf} project\footnote{\url{https://github.com/autonomousvision/graf}}. Images are rendered via only the ``coarse'' network of the decoder, \ie, removing the hierarchical sampling. The sampling range of the scaling factor $\alpha$ of Eq.~\eqref{eq:ray_sample} is initialized as [0.9, 1.0], where the lower bound is exponentially annealed to 0.06 during training. Encoder is initialized with the ImageNet-1K~\cite{deng2009imagenet} pretrained weights, decoder is initialized with $\pi$-GAN pretrained weights, while the discriminator is randomly initialized.
We freeze the parameters of the decoder, and set the learning rate of the encoder and discriminator to $1\times10^{-4}$ and $2\times10^{-5}$, respectively. Ranger optimizer\footnote{\url{https://github.com/lessw2020/Ranger-Deep-Learning-Optimizer}} is used for both encoder and discriminator. We set the training batch size to 8, and use V100 GPUs to train all related models for 200,000 iterations.

\textbf{Inference.} To render qualitative results, rays are cast with size $512\times512$ and depth step 72 in the inference. Besides, ``fine'' network is activated to enable hierarchical sampling. For GAN-inversion used by pix2pixHD as mentioned in Section~\ref{sec:exp_set}, we adopt the implementation from the $\pi$-GAN project\footnote{\url{https://github.com/marcoamonteiro/pi-GAN}}, and set the iteration number to 700 as suggested.

\section{Additional Visual Results} \label{sec:result}
 
In the following three pages, we will present additional visual results for the proposed Sem2NeRF regarding free-viewpoint image generation (see Section~\ref{sec:supp_free}), semantic mask editing (see Section~\ref{sec:supp_edit}) and multi-modal synthesis (see Section~\ref{sec:supp_style}). Results are demonstrated on both CelebAMask-HQ and CatMask, and they are all best viewed in high quality color image.

\clearpage

\subsection{Free-viewpoint Image Generation} \label{sec:supp_free}

\begin{figure}[h!]
    \centering
    \includegraphics[width=0.99\textwidth]{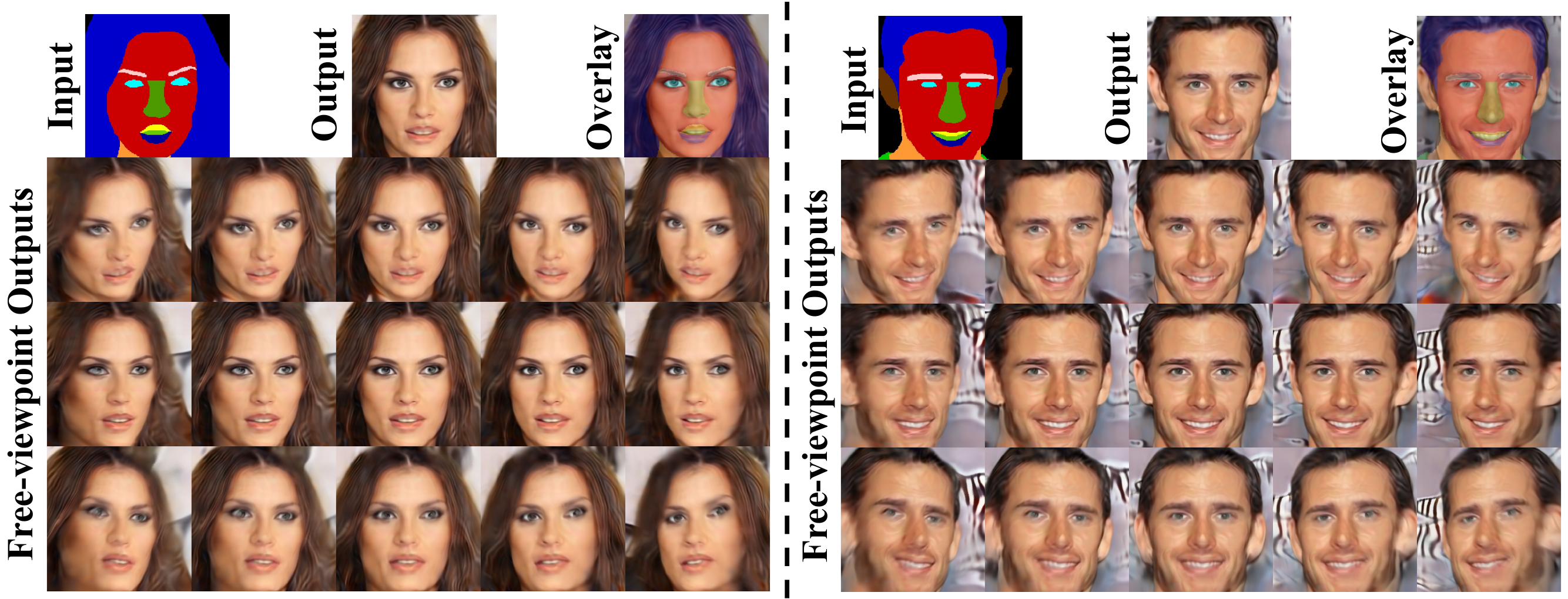}
    \caption{Free-viewpoint image generation on CelebAMask-HQ. ``Output'' refers to the generated image that has the same viewing direction as the ``Input'', and ``Overlay'' shows the results of overlaying ``Output'' with ``Input'', so as to better demonstrate the mapping accuracy}
    \label{fig:supp_free_celeba}
\end{figure}

\begin{figure}[h!]
    \centering
    \includegraphics[width=0.99\textwidth]{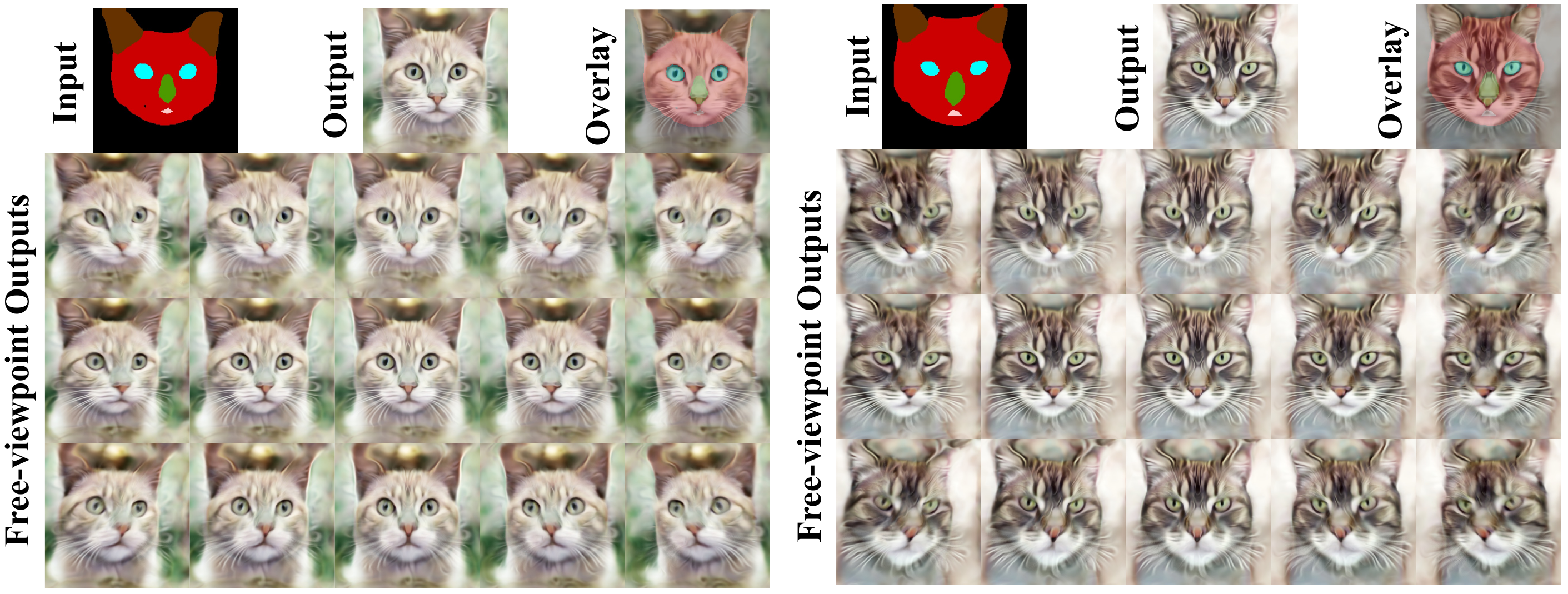}
    \caption{Free-viewpoint image generation on CatMask. ``Output'' refers to the generated image that has the same viewing direction as the ``Input'', and ``Overlay'' shows the results of overlaying ``Output'' with ``Input'', so as to better demonstrate the mapping accuracy}
    \label{fig:supp_free_cat}
\end{figure}

Additional visual results of free-viewpoint image generation on CelebAMask-HQ and CatMask datasets are given in Figure~\ref{fig:supp_free_celeba} and Figure~\ref{fig:supp_free_cat}, respectively. 

\clearpage

\subsection{Semantic Mask Editing} \label{sec:supp_edit}

\begin{figure}[h!]
    \centering
    \includegraphics[width=0.99\textwidth]{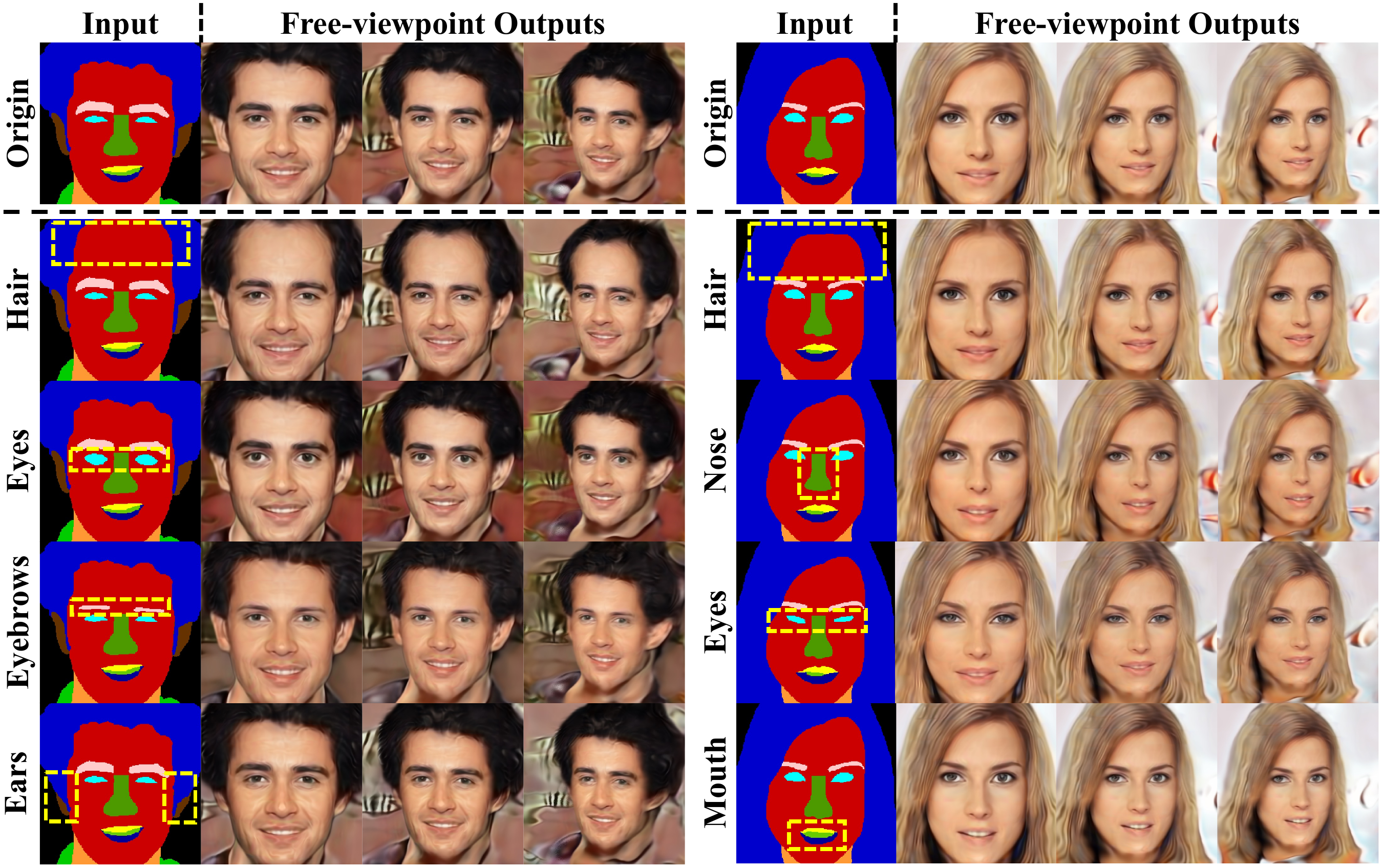}
    \vspace{-3mm}
    \caption{Semantic mask editing on CelebAMask-HQ. The first row shows the results of the original semantic masks, while the following rows give the results of editing the mentioned area, highlighted with yellow-dash box. Three viewpoints are given for each group, with the first one having the same viewing direction as the input}
    \vspace{-3mm}
    \label{fig:supp_edit_celeba}
\end{figure}

\begin{figure}[h!]
    \centering
    \includegraphics[width=0.99\textwidth]{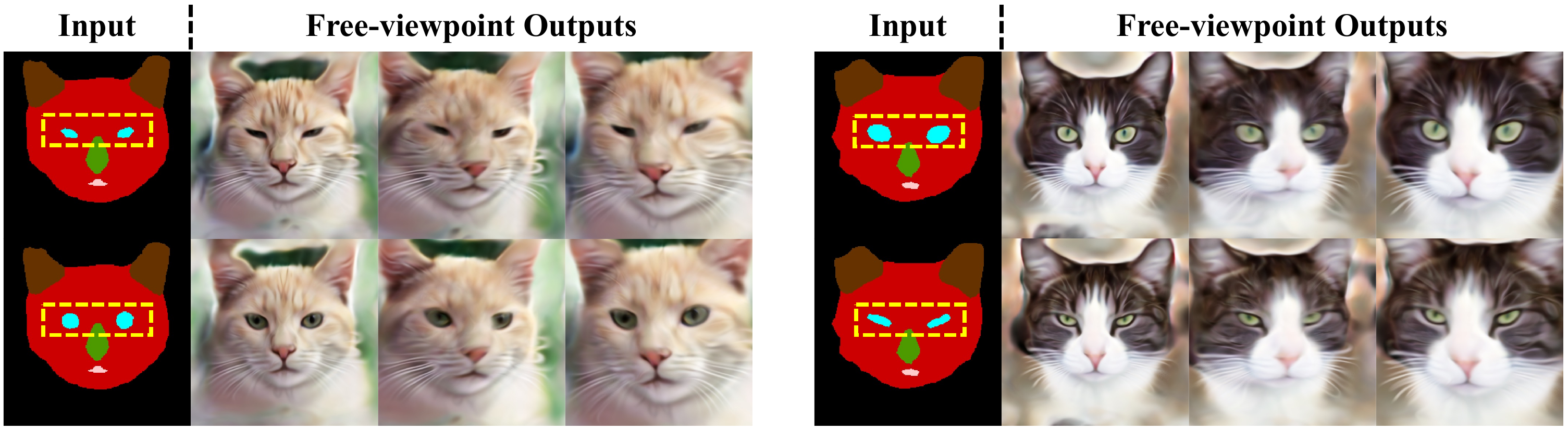}
    \vspace{-3mm}
    \caption{Semantic mask editing on CatMask. Edited regions are highlighted with yellow-dash box. The first viewpoint has the same pose as the input}
    \vspace{-3mm}
    \label{fig:supp_edit_cat}
\end{figure}

Additional visual results of semantic mask editing on CelebAMask-HQ and CatMask datasets are given in Figure~\ref{fig:supp_edit_celeba} and Figure~\ref{fig:supp_edit_cat}, respectively. 

\subsection{Multi-modal Synthesis} \label{sec:supp_style}

\begin{figure}[t!]
    \centering
    \includegraphics[width=0.99\textwidth]{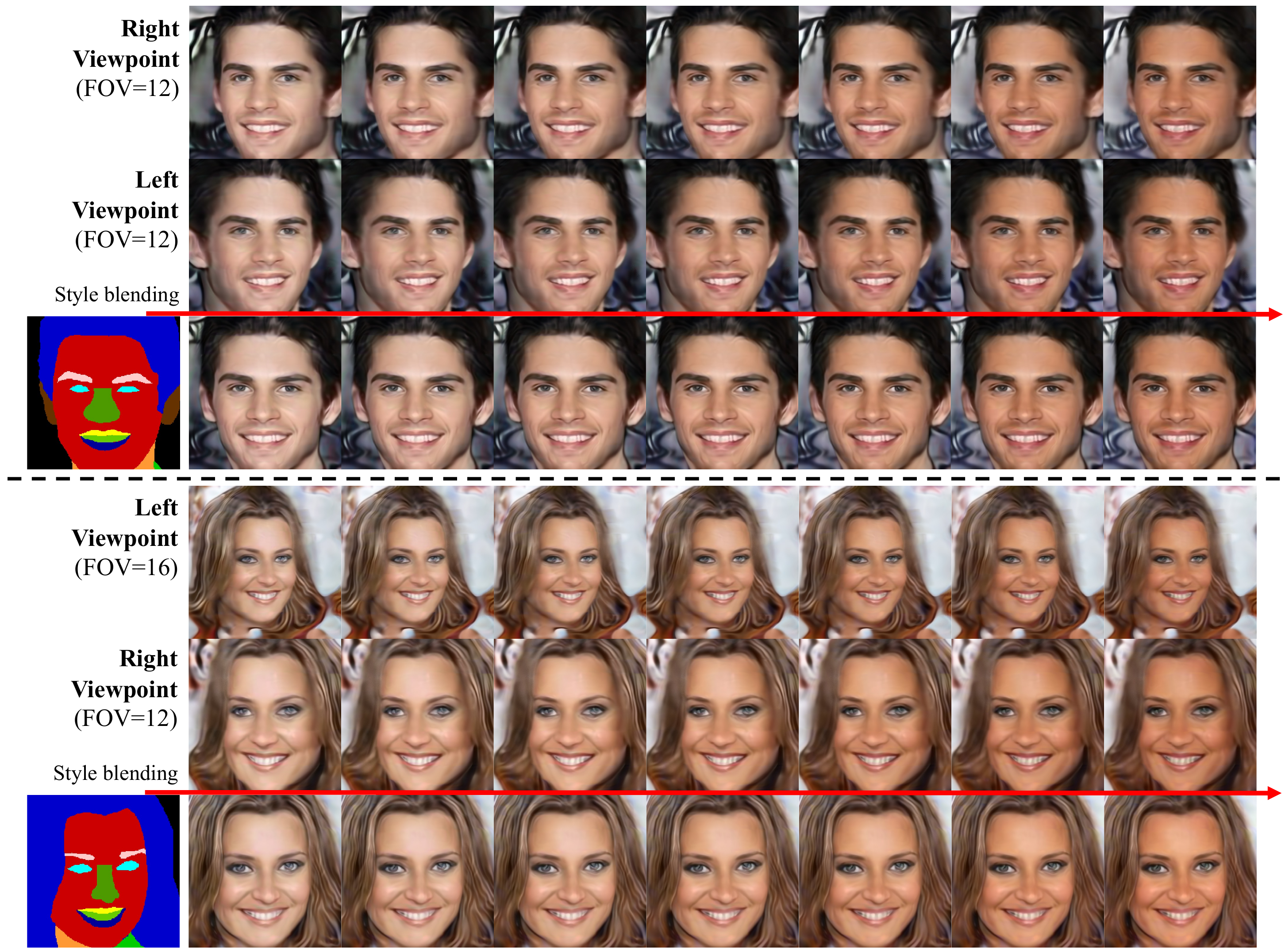}
    \vspace{-3mm}
    \caption{Multi-modal synthesis on CelebAMask-HQ. Styles are linearly blended from left to right. The last viewpoint in each group has the same pose as the input}
    \vspace{-3mm}
    \label{fig:supp_style_celeba}
\end{figure}

\begin{figure}[t!]
    \centering
    \includegraphics[width=0.99\textwidth]{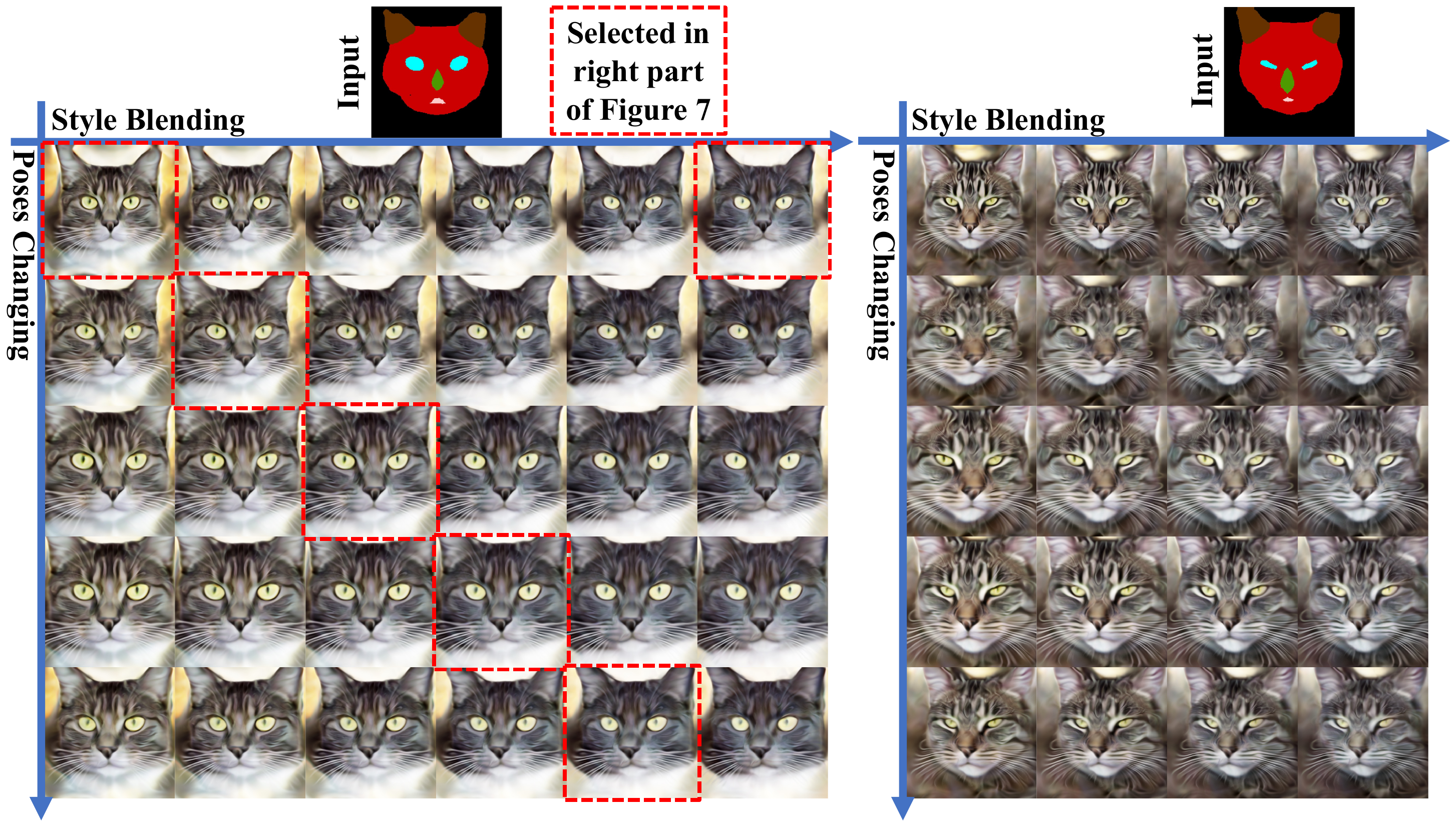}
    \vspace{-3mm}
    \caption{Multi-modal synthesis on CatMask. Left case shows the full version of Figure~\ref{fig:cat_cases} (right part), where the selected images are highlighted in red-dash border. Images in the first row have the same viewing direction as the input}
    \vspace{-3mm}
    \label{fig:supp_style_cat}
\end{figure}

Additional visual results regarding multi-modal synthesis on CelebAMask-HQ and CatMask datasets are given in Figure~\ref{fig:supp_style_celeba} and Figure~\ref{fig:supp_style_cat}, respectively.

\end{document}